\documentclass{article} % For LaTeX2e

\usepackage{iclr2026_arxiv,times}

\usepackage{amsmath,amsfonts,bm}

\def\eqref#1{equation~\ref{#1}}
\def\1{\bm{1}}

\DeclareMathAlphabet{\mathsfit}{\encodingdefault}{\sfdefault}{m}{sl}
\SetMathAlphabet{\mathsfit}{bold}{\encodingdefault}{\sfdefault}{bx}{n}

\DeclareMathOperator*{\argmax}{arg\,max}

\usepackage{hyperref}
\usepackage{url}
\usepackage{graphicx}
\usepackage{enumitem}
\usepackage{bbding}
\usepackage{amsmath}
\usepackage{amssymb}
\usepackage{booktabs}
\usepackage{multirow}
\usepackage{wrapfig}
\usepackage{xcolor}
\usepackage{listings}
\usepackage{color}
\usepackage{colortbl}
\usepackage{tcolorbox}
\usepackage{wrapfig}
\usepackage[utf8]{inputenc}
\usepackage{subcaption}

\usepackage{graphicx}
\usepackage{enumitem}
\usepackage{multirow}
\usepackage{colortbl}
\usepackage{booktabs}
\usepackage{amsmath}
\usepackage{amssymb}
\usepackage{xspace}
\usepackage{soul}
\usepackage{comment}
\usepackage{epigraph}
\usepackage{algorithm}
\usepackage{algpseudocode}
\usepackage{tcolorbox}
\usepackage{pifont}

\usepackage{caption}
\usepackage{hyperref}
\usepackage{url}
\usepackage{cleveref}

\definecolor{Gray}{gray}{0.90}
\definecolor{skyblue}{rgb}{0.04,0.40,0.80}
\hypersetup{
    colorlinks,
    linkcolor={red},
    citecolor={skyblue},
    urlcolor={magenta}
}

\definecolor{forestgreen}{rgb}{0.13,0.55,0.13}

\newcommand{\up}[1]{\textcolor{red}{\,$\uparrow$#1}}
\newcommand{\down}[1]{\textcolor{forestgreen}{\,$\downarrow$#1}}

\title{MomentSeg: Moment-Centric Sampling for Enhanced Video Pixel Understanding}

\author{Ming Dai, Sen Yang, Boqiang Duan, Wankou Yang, Jingdong Wang}

\author{Ming Dai$^{1,2*}$ \hspace{1em}
Sen Yang$^{2*}$ \hspace{1em}
Boqiang Duan$^{2}$ \hspace{1em}
Wankou Yang$^{1}$ \hspace{1em}
Jingdong Wang$^{2}$ \\
[1.5ex]
$^{1}$Southeast University\qquad
$^{2}$Baidu VIS
}

\iclrfinalcopy % Uncomment for camera-ready version, but NOT for submission.
\begin{document}

\maketitle

\renewcommand{\thefootnote}{*}
\footnotetext{This work was done during an internship at Baidu VIS.}
\renewcommand{\thefootnote}{\arabic{footnote}}

\begin{figure}[h]
  \centering
  \vspace{-4mm}
  \includegraphics[width=0.9\linewidth]{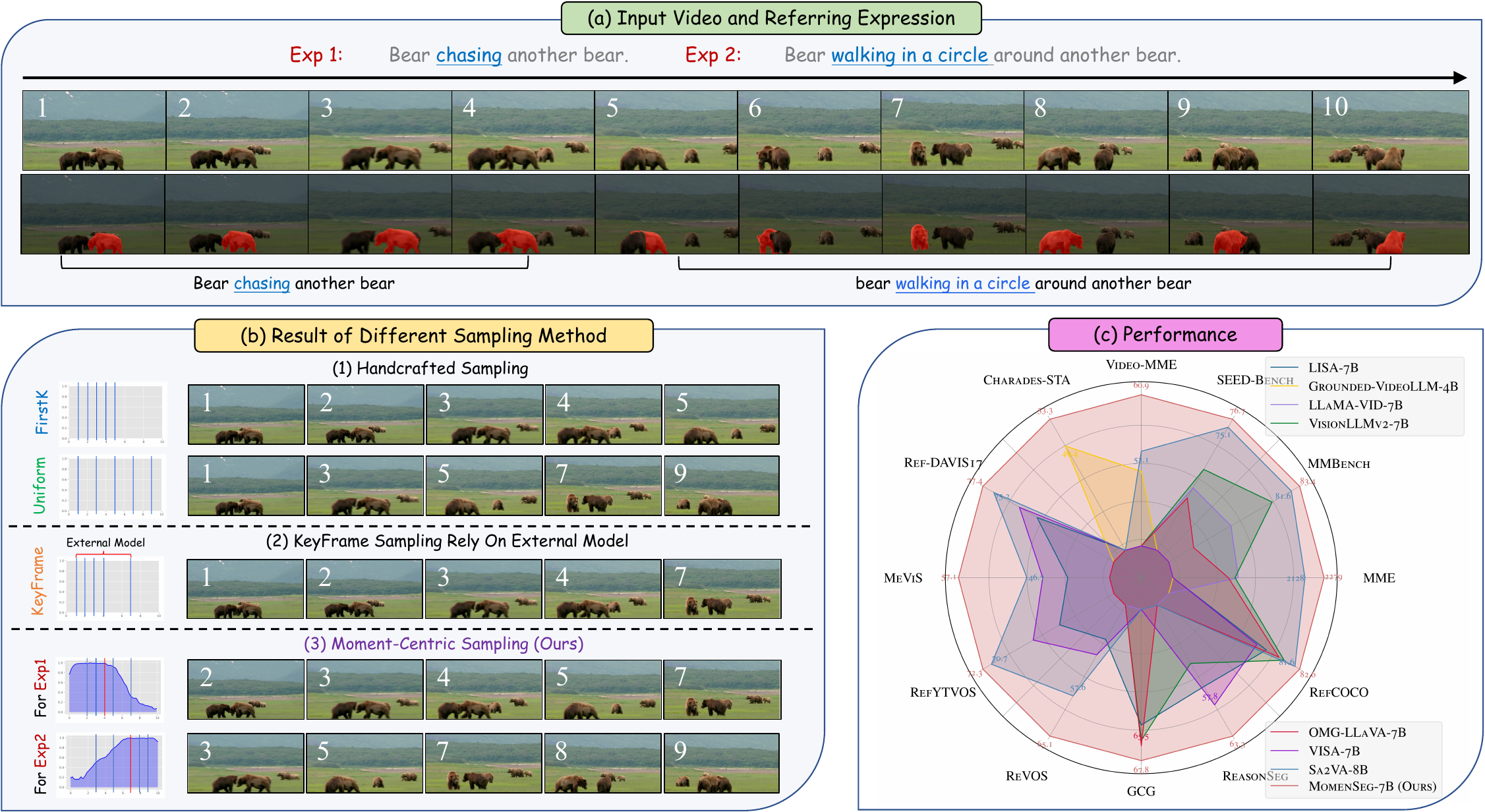}
  % \vspace{0mm}
  \caption{\textbf{Illustration of Moment-Centric Sampling.} 
  (a) Two example expressions with 10 selected video frames and GT masks. 
  (b) Comparison of sampling strategies: handcrafted, external keyframe–based, and our proposed \emph{Moment-Centric Sampling} (MCS). 
  MCS performs dense sampling at critical moments and sparse sampling elsewhere, without relying on external keyframe models. 
  (c) MomentSeg achieves superior performance across RefVOS, TSG, and QA tasks compared with other LMM-based methods.}
  \label{fig:teaser}
  % \vspace{-2mm}
\end{figure}

\begin{abstract}
Referring Video Object Segmentation (RefVOS) seeks to segment target objects in videos guided by natural language descriptions, demanding both temporal reasoning and fine-grained visual comprehension. Existing sampling strategies for LLM-based approaches typically rely on either handcrafted heuristics or external keyframe models. The former often overlooks essential temporal cues, while the latter increases system complexity. To address this, we propose a unified framework that jointly optimizes Temporal Sentence Grounding (TSG) and RefVOS, naturally incorporating key moment grounding capability. During training, we introduce a novel TSG paradigm that employs a dedicated \texttt{[FIND]} token for key moment identification through temporal token similarity matching, thereby avoiding the need for external timestamp encodings. For inference, we design a Moment-Centric Sampling (MCS) strategy that densely samples informative moments while sparsely sampling non-essential frames, preserving both motion details and global context. To further enhance tracking stability, we develop Bidirectional Anchor-updated Propagation (BAP), which leverages the most relevant moment as start point for high-quality mask initialization and dynamically updates at sampled points to mitigate accumulated errors. 
% Extensive experiments demonstrate that MomentSeg achieves the SOTA performance across multiple benchmarks.
Code and model will be available at: \url{https://github.com/Dmmm1997/MomentSeg}
\end{abstract}

\section{Introduction}
\label{sec:intro}

% Motivation
RefVOS has attracted considerable attention in recent years. The task involves localizing and segmenting target objects at the pixel level in videos based on natural language expressions~\citep{gavrilyuk2018actor, khoreva2019video, seo2020urvos}. It requires models to interpret temporal action semantics described in text and to maintain consistent tracking of the referred object across frames. Recent Transformer-based methods~\citep{mttr, SOC, DsHmp, pan2025semantic} mainly focus on improving temporal reasoning and inter-frame consistency. In contrast, large multimodal models (LMMs)~\citep{Qwen2.5-VL,chen2024internvl, li2024survey}, pre-trained on massive video-text corpora, demonstrate stronger potential for advancing pixel-level video understanding tasks~\citep{munasinghe2024videoglamm, lin2025glus, yuan2025sa2va}.

A central challenge in RefVOS lies in effectively sampling keyframes that are relevant to the linguistic expression. Given the variable and often large number of frames, feeding all frames into the model is both computationally prohibitive and redundant. Consequently, identifying and selecting semantically meaningful keyframes is worth exploring.
Existing approaches employ diverse strategies for keyframe selection, as summarized in Table~\ref{tab:formulation}. For instance, Sa2VA~\citep{yuan2025sa2va} simply selects the first K frames, disregarding content-specific cues such as object presence or action timing. VideoLISA~\citep{bai2024one} compresses temporal information and samples frames uniformly to retain coarse timing information but still overlooks referentially critical frames. In contrast, methods such as VISA~\citep{yan2024visa}, ViLLA~\citep{zheng2024villa}, VRS-HQ~\citep{gong2025devil}, and GLUS~\citep{lin2025glus} depend on separately trained models to pinpoint the most relevant timestamps, then either use tracking networks~\citep{cheng2022xmem} to generate mask trajectories or select key sampling frames to guide large-model segmentation.
 While these methods enable the identification of keyframes, they are dependent on these additional models.

\begin{figure}
  \centering
  \vspace{-8mm}
  \includegraphics[width=\linewidth]{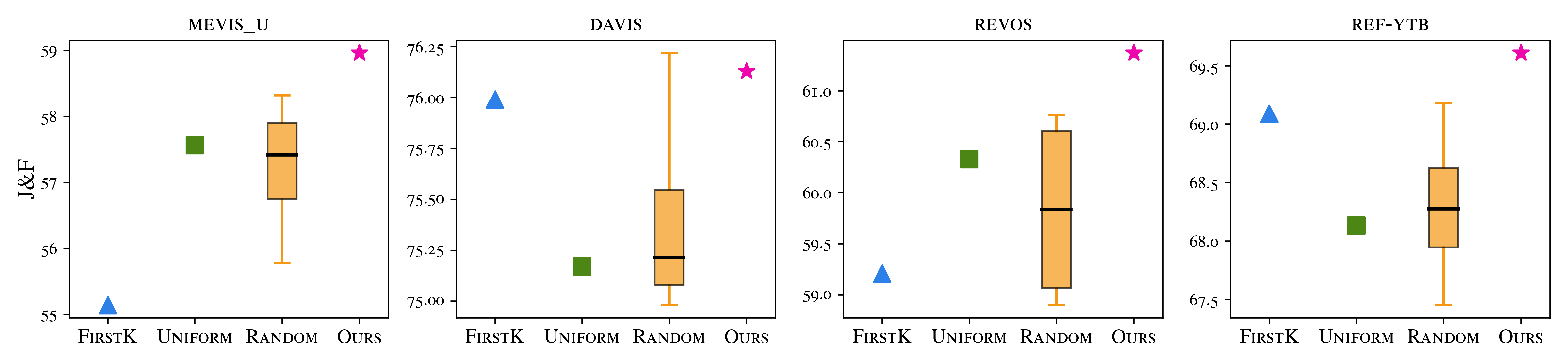}
  \vspace{-6mm}
  \caption{\textbf{Impact of Sampling Strategies.} 
  We evaluate various sampling strategies for the RefVOS task on four datasets. 
  Five independent runs of random sampling show high variance, highlighting the need for a robust sampling mechanism. 
  Our proposed MCS consistently outperforms alternative methods across all datasets.}
  \label{fig:sample_std}
  \vspace{-5mm}
\end{figure}

Building on prior explorations, we first pose a fundamental question: \textbf{\emph{Is keyframe selection necessary?}}
From an \textbf{\textit{intuitive}} perspective, failing to capture critical moments in a video risks missing the target object's appearance or the described action, which directly hinders accurate segmentation. Conversely, precisely identifying and leveraging text-relevant key moments mitigates the misleading effects of irrelevant frames on LMMs while reducing computational redundancy.
From a \textbf{\textit{quantitative}} perspective, as shown in Fig.~\ref{fig:sample_std}, we evaluate several sampling strategies across multiple RefVOS datasets. Two consistent patterns emerge: \textbf{(1)} random sampling exhibits high variance, highlighting the strong influence of sampling strategy; and \textbf{(2)} for scenarios involving motion-driven (e.g., MeVIS~\citep{ding2023mevis}) and complex semantic understanding (e.g., ReVOS~\citep{yan2024visa}), uniform sampling often surpasses the firstK approach. In contrast, firstK performs better on datasets~\citep{khoreva2019video,seo2020urvos} where targets appear early in the video and descriptions rely less on temporal cues. Likewise, MeVIS, which includes numerous action-centric descriptions and scenes with multiple visually similar objects, poses a particular challenge: if sampled frames miss the described action while static cues match several candidates, the model struggles to disambiguate the target. This analysis underscores that \textit{keyframe selection is critical for RefVOS}.

Next, we consider the following question: \textbf{\emph{How can we effectively identify keyframes, and how can these keyframes enhance RefVOS performance?}} Prior work offers two primary strategies: \textbf{(1)} Keyframe for tracking initialization. VISA~\citep{yan2024visa} employs LLAMA-VID~\citep{li2025llama} to select the most critical frame as the tracking initialization timestamp, and applies XMem~\citep{cheng2022xmem} for subsequent tracking. \textbf{(2)} Keyframes for LMM understanding. ViLLA~\citep{zheng2024villa} leverages Grounded-VideoLLM~\citep{wang2024grounded} to generate relevant temporal intervals and then samples frames inside and outside these intervals using handcrafted ratios, subsequently feeding these frames to LMM. VRS-HQ~\citep{gong2025devil} further combines CLIP-based~\citep{clip2021} vision-language similarity with SAM2 confidence scores to filter keyframes before passing them to the LMM.
While effective, all these methods depend on external models for keyframe identification, inevitably increasing overall system complexity.

\begin{figure}[t]
  \centering
  \vspace{-3mm}
  \includegraphics[width=1.0\linewidth]{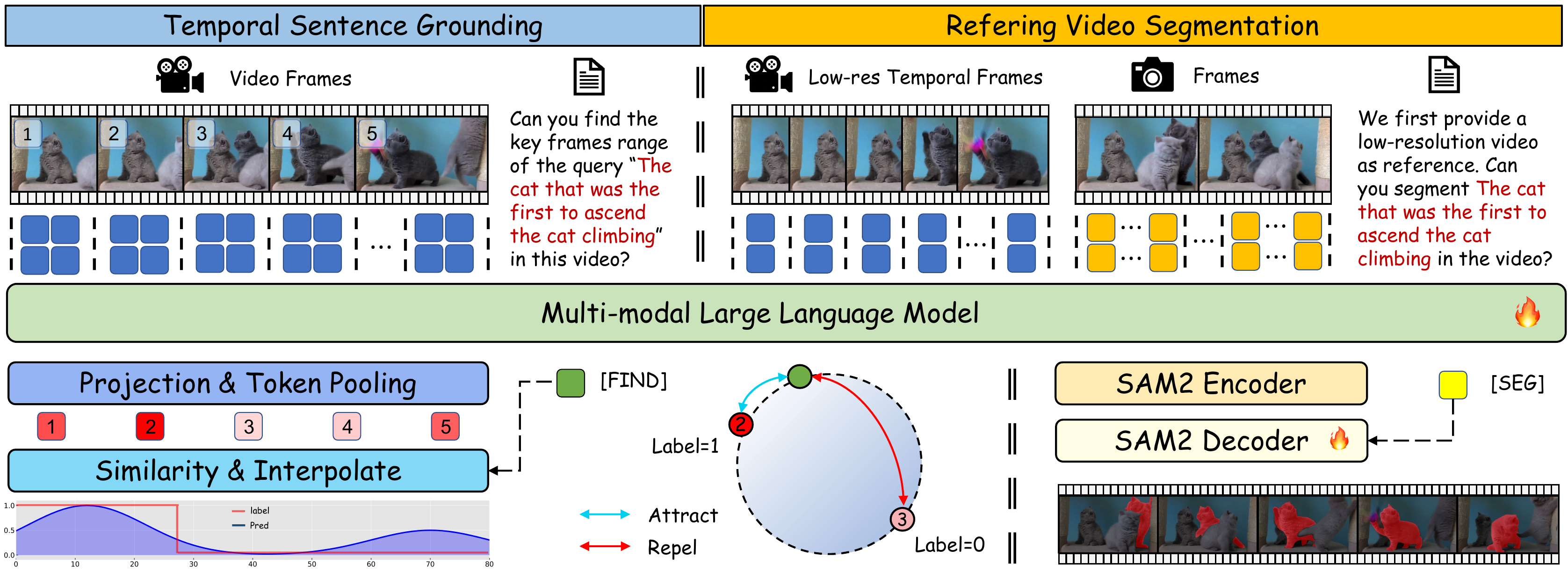}
  \vspace{-5mm}
   \caption{\textbf{Training framework of the proposed MomentSeg model}. In the TSG paradigm, we employ the Qwen2.5-VL~\citep{Qwen2.5-VL} video encoder with low-resolution image inputs. We further introduce a \texttt{[FIND]} token trained under a contrastive learning scheme. In the RefVOS paradigm, both low-resolution uniformly sampled frames and high-resolution continuously sampled frames are used as inputs. Supervision is provided via the \texttt{[SEG]} token, with segmentation masks generated by the SAM2 decoder.}
  \label{fig:train_framework}
  \vspace{-3mm}
\end{figure}

In this paper, we propose a unified framework for the synchronized optimization of TSG and RefVOS. \textbf{\textit{During training}}, as shown in Fig.~\ref{fig:train_framework}, we introduce a special \texttt{[FIND]} token for the TSG task, similar to the \texttt{[SEG]} token in RefVOS. This token computes similarity with sampled video frame tokens, offering two key advantages: \textbf{First}, for RefVOS, it eliminates the need for external models for keyframe selection, as the model natively identifies key moments. \textbf{Second}, it obviates the need for explicit timestamp encoding~\citep{chen2024timemarker} by leveraging a frame-level matching mechanism between \texttt{[FIND]} and temporal tokens.
\textbf{\textit{During inference}}, as depicted in Fig.~\ref{fig:inference_framework}, we propose a Moment-Centric Sampling (MCS). MCS leverages the similarity distribution from the \texttt{[FIND]} token to densely sample video moments highly relevant to the description, while sparsely sampling less critical frames. This approach effectively preserves crucial motion cues while maintaining a global temporal context. For the RefVOS task, consistent with Sa2VA~\citep{yuan2025sa2va}, a single \texttt{[SEG]} token is used to represent the referential object throughout the video. We further introduce a novel Bidirectional Anchor-updated Propagation (BAP) strategy, which is composed of two key components:
\textbf{(1)} \textit{Bidirectional Propagation}: The central keyframes identified by MCS serve as starting points for propagation in both forward and backward temporal directions.
\textbf{(2)} \textit{Anchor-updated}: At these sampled time points, the target mask is updated and the prior memory is adaptively refreshed, which mitigates cumulative tracking errors.

\noindent In summary, our key contributions are as follows:
\textbf{(1)} We propose \textbf{MomentSeg}, a unified framework that integrates temporal sentence grounding (TSG) and referring video object segmentation (RefVOS). For TSG, MomentSeg obviates explicit timestamp encoding by leveraging the similarity between the \texttt{[FIND]} token and temporal tokens. For RefVOS, our model inherently identifies text-relevant keyframes, thereby removing the dependency on external keyframe-selection models.
\textbf{(2)} We introduce \textbf{Moment-Centric Sampling (MCS)}, a similarity-driven sampling strategy that densely samples frames around relevant moments and sparsely samples non-essential frames. MCS preserves salient motion cues while retaining global temporal context.
\textbf{(3)} We present \textbf{Bidirectional Anchor-updated Propagation (BAP)}, which initiates mask propagation from query-relevant keyframes and adaptively updates the target mask at sampled temporal points. BAP enhances tracking robustness and mitigates cumulative propagation errors.
\textbf{(4)} MomentSeg achieves new SOTA performance, delivering a 5\% improvement on the MeVIS and a 6\% gain on the ReVOS.

\section{Related Work}
\label{sec:related}

\noindent \textbf{Referring Video Object Segmentation.}
RefVOS aims to segment targets in a video conditioned on a given description. Existing approaches typically employ object queries to fuse expressions with visual features for referent identification~\citep{wu2022referformer, wu2023onlinerefer, SOC, tang2023temporal, yuan2024losh}, while a line of work emphasizes motion aggregation to capture dynamic cues~\citep{ding2023mevis, DsHmp, fang2025decoupled}. Concurrently, LMMs~\citep{liu2024llava} have been used to reason over compositional and complex expressions~\citep{zhu2023tracking, yan2024visa, bai2024one, munasinghe2024videoglamm, lin2025glus,groundmore}. To address these gaps, we propose \textbf{MomentSeg}, which performs keyframe selection natively, removing the need for external modules. We further introduce Bi-directional Anchor-updated Propagation (BAP), which propagates masks forward and backward from key moments and adaptively updates memory at sampled nodes, enhancing segmentation robustness and reducing cumulative errors.

\begin{table}[t]
\centering
\vspace{-3mm}
\caption{Comparison of \emph{keyframe models}, \emph{sampling}, and \emph{information} utilized in existing RefVOS MLLMs. ``cont.'' means continuous sampling. Unlike prior methods, our approach inherently supports keyframe selection while simultaneously leveraging both ``Global'' and ``Essential'' information during inference.}
\vspace{-3mm}
\resizebox{1.0\textwidth}{!}{
    \begin{tabular}{l|l|cc|cc|c}
        \toprule
        \multirow{2}{*}{Method} & \multirow{2}{*}{External Keyframe Model}  &\multicolumn{2}{c|}{Training} & \multicolumn{2}{c|}{Inference} & \multirow{2}{*}{\texttt{[SEG]}}\\ 
        &  & Sampling & Information & Sampling & Information & \\
        \midrule
        VideoLISA~\citep{bai2024one} &  / & Uniform & Global & Uniform  & Global & 1  \\
        VISA~\citep{yan2024visa} & LLaMA-VID~\citep{li2025llama} & Random & Random & Uniform + Cont. & Global + Local & 1  \\
        ViLLa~\citep{zheng2024villa} & G-VideoLLM~\citep{wang2024grounded} & Uniform & Global & Cont. & Local & $N$  \\
        GLUS~\citep{lin2025glus} & Chat-UniVi~\citep{jin2024chat}  & Uniform + Cont. & Global + Local & Uniform + Cont. & Global + Local & $N$ \\ 
        Sa2VA~\citep{yuan2025sa2va} & / & Random & Random & FirstK & Local & 1  \\
        VRS-HQ~\citep{gong2025devil} & CLIP~\citep{clip2021} & Uniform & Global & Selection & Essential & $N$ \\ 
        \midrule
        \textbf{MomentSeg (Ours)} & Inherently & Uniform + Cont. & Global + Local & Selection & Global + Essential & 1  \\
        \bottomrule
    \end{tabular}}
    
    \vspace{-3mm}
    \label{tab:formulation}
\end{table}

\noindent \textbf{LLM-based RefVOS Sampling Strategies.} Existing LLM-based RefVOS methods adopt diverse frame-sampling and keyframe-selection schemes (Table~\ref{tab:formulation}). VideoLISA~\citep{bai2024one} and Sa2VA~\citep{yuan2025sa2va} adopt handcrafted sampling strategies, using uniform or firstK sampling to generate a global \texttt{[SEG]} token. In contrast, VISA~\citep{yan2024visa}, ViLLA~\citep{zheng2024villa}, and GLUS~\citep{lin2025glus} rely on external models (e.g., LLaMA-VID~\citep{li2025llama}, Grounded-VideoLLM~\citep{wang2024grounded}, Chat-UniVi~\citep{jin2024chat}) for keyframe selection, while VRS-HQ~\citep{gong2025devil} selects keyframes via CLIP-based vision–language similarity~\citep{clip2021}. Such dependencies increase pipeline complexity and limit adaptability. By comparison, \textbf{MomentSeg} performs keyframe selection natively and introduces Moment-Centric Sampling (MCS), which applies dense sampling around text-relevant moments and sparse sampling elsewhere, preserving salient motion cues while maintaining a compact global temporal context.

\noindent \textbf{Timestamps Encoding in Temporal Sentence Grounding.} TSG~\citep{gao2017tall, caba2015activitynet, zala2023hierarchical, wang2024grounded,li2025llava,liu2025videomind} aims to localize a video segment corresponding to a language query. A key challenge for LMM-based methods lies in developing an effective timestamp encoding strategy. Approaches such as TimeChat~\citep{ren2024timechat}, VTimeLLM~\citep{huang2024vtimellm}, and LITA~\citep{lita} leverage instruction tuning for plain text temporal grounding. Momentor~\citep{momentor} injects explicit temporal position encodings into frame-level features to improve temporal localization, while VTG-LLM~\citep{vtgllm} incorporates absolute-time tokens. Vid2Seq~\citep{vid2seq} and Grounded-VideoLLM~\citep{wang2024grounded} adopt discrete or relative temporal tokens to circumvent direct timestamp encoding. NumberIT~\citep{numberit} injects sequential frame indices into images for grounding. In contrast, \textbf{MomentSeg} avoids timestamp encoding by employing a specialized \texttt{[FIND]} token that performs frame-level temporal matching.

\section{Methods}

\subsection{Preliminaries}
\label{sec:prelim}
\noindent \textbf{Referring Video Object Segmentation:} 
Given an input video with $T$ frames $I_{1:T} \in \mathbb{R}^{T\times H\times W\times 3}$ and a referring language expression $R$, RefVOS aims to train a model $\phi_\theta$ that predicts binary segmentation masks ${M}_{1:T}$ for the referred object:
\begin{equation}\label{eqn:refvos}
   \setlength{\abovedisplayskip}{2pt}
   \setlength{\belowdisplayskip}{2pt}
   M_{1:T} = \phi_\theta(I_{1:T}, R).
\end{equation}

\noindent \textbf{LMMs-based RefVOS Paradigm:}  
Recent LMM-based RefVOS approaches build upon image-level referring segmentation models such as LISA~\citep{lai2024lisa}, which employ a dedicated \texttt{[SEG]} token to represent the target and a segmentation decoder $\mathtt{Dec}$ to generate masks:
\begin{equation}\label{eqn:lisa}
    \setlength{\abovedisplayskip}{2pt}
    \setlength{\belowdisplayskip}{2pt}
    \texttt{[SEG]} = \mathtt{LMM}(I, R), \ \ \ M = \mathtt{Dec}(I, \texttt{[SEG]}).
\end{equation}
This paradigm extends to videos by adopting a multi-image formulation, where the LMM processes multiple frames $I_{1:N}$ and segmentation tokens decode masks for each frame:
\begin{equation}\label{eqn:vos}
   \setlength{\abovedisplayskip}{2pt}
   \setlength{\belowdisplayskip}{2pt}
   \texttt{[SEG]} = \mathtt{LMM}(I_{1:N}, R), \ \ \ M_{1:N} = \mathtt{Dec}(I_{1:N}, \texttt{[SEG]}).
\end{equation}
This equation exemplifies a ``One Token Seg All'' paradigm~\citep{bai2024one}, in which a single \texttt{[SEG]} token represents the referential object throughout the video. Here, $N$ denotes the number of frames provided to the LMM, which is typically smaller than the full sequence length $T$.

\subsection{Model Architecture of MomentSeg}
\label{subsec:model_architecture}

\subsubsection{Overall Pipeline}
\label{subsubsec:overall_pipeline}

\noindent \textbf{Training Pipeline:}
Fig.~\ref{fig:train_framework} presents the training paradigm of MomentSeg for both TSG and RefVOS. For RefVOS, we follow a scheme similar to Sa2VA~\citep{yuan2025sa2va}, while additionally incorporating low-resolution, uniformly sampled frames $I^l_{1:L}$ and high-resolution, densely sampled frames $I^h_{1:N}$ from temporal segments where the referential object is present. The sampling of these low-resolution frames is referred to as \emph{Temporal Token Injecting} in this paper. Notably, the number of tokens introduced into the LMM remains minimal, at no more than 16 tokens per frame in our setting. The low-resolution frames $I^l_{1:L}$ supply global temporal context, whereas the high-resolution frames $I^h_{1:N}$ capture fine-grained spatial details. A SAM2 decoder is then applied to the high-resolution frames and is supervised as follows:
\begin{equation}\label{eqn:momentseg_refvos}
   \setlength{\abovedisplayskip}{2pt}
   \setlength{\belowdisplayskip}{2pt}
   \texttt{[SEG]}^h = \mathtt{LMM}(I^l_{1:L}, I^h_{1:N}, R), \quad
   M_{1:N} = \mathtt{Dec}(I^h_{1:N}, \texttt{[SEG]}^h).
\end{equation}
For TSG, we introduce a special token \texttt{[FIND]} to identify relevant frames through similarity matching with temporal features. This design removes the need for external timestamp encoding, as temporal correspondence is established directly via frame-level matching. Specifically, we first extract the temporal tokens corresponding to the LMM output. These tokens are projected into a feature space using an MLP layer and subsequently average-pooled to obtain frame representations $T^v \in \mathbb{R}^{L_t \times C}$. The \texttt{[FIND]} tokens, $T^f \in \mathbb{R}^{N_f \times C}$ (where $N_f$ denotes the number of \texttt{[FIND]} tokens in a sample under the multi-round dialogue paradigm), are mapped through the same MLP layer. We then compute the similarity matrix $\ell \in \mathbb{R}^{N_f \times L_t}$ between frame tokens and \texttt{[FIND]} tokens to identify text-relevant frames. The matching loss is defined as:
\begin{equation}\label{eqn:find_loss}
\setlength{\abovedisplayskip}{2pt}
\setlength{\belowdisplayskip}{2pt}
\begin{aligned}
\ell_{ij} = \frac{(T^f_i)^\top T^v_j}{\|T^f_i\| \, \|T^v_j\| \, \tau},
\mathcal{L}_{find} = \frac{1}{|\Omega|} \sum_{(i,j)\in \Omega}
\Big[-\lambda_{p}y_{ij}\log\sigma(\ell_{ij})-(1-y_{ij})\log(1-\sigma(\ell_{ij}))\Big],
\end{aligned}
\end{equation}
where $\sigma$ denotes the sigmoid function, $T^v_i \in \mathbb{R}^C$ is the $i$-th pooled frame token, $T^f_j \in \mathbb{R}^C$ is the $j$-th \texttt{[FIND]} token, $y_{ij} \in \{0,1\}$ is the match label, $\Omega$ is the set of valid pairs, and $\tau$ is the temperature factor, set to 0.07. The positive sample weight $\lambda_{p}$ is set to 2.0. 
Notably, since Qwen2.5-VL compresses the temporal dimension during video encoding (with $L_t$ being half of $L$), we apply bilinear interpolation to the similarity distribution to ensure alignment with the original input length.

\begin{figure}[t]
  \centering
  \vspace{-3mm}
  \includegraphics[width=1.0\linewidth]{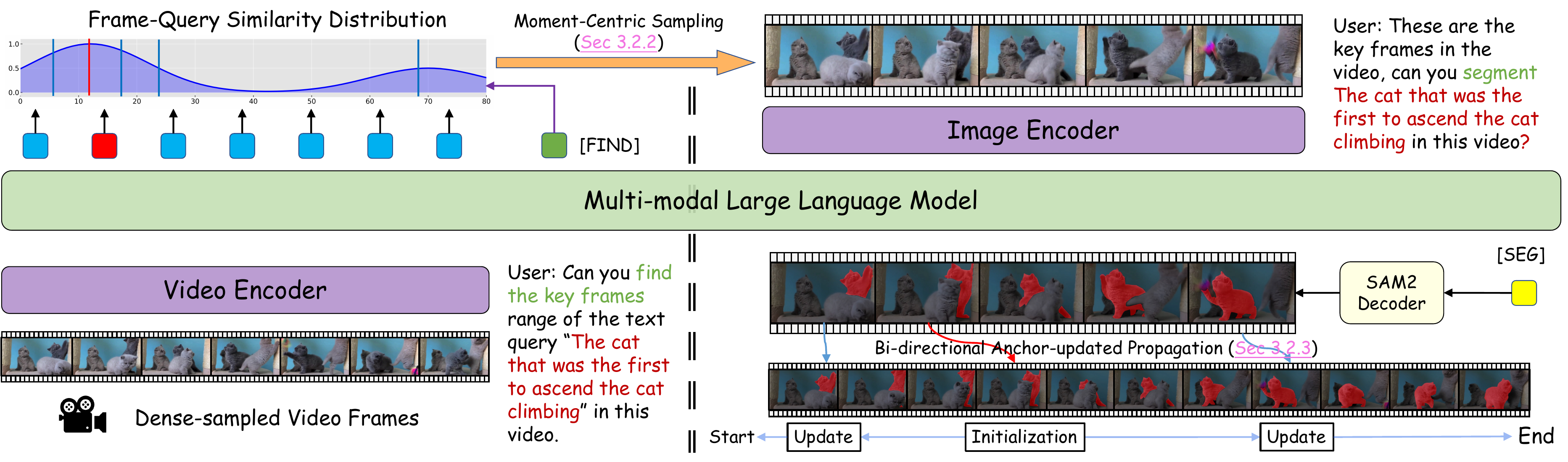}
  \vspace{-5mm}
  \caption{\textbf{Inference framework of the proposed MomentSeg model.} First, we use dense-sampled video frames to find the frame-query similarity distribution related to the description, then apply a Moment-Centric Sampling (MCS) to select key frames from the sequence. These key frames are input to the model, along with the RefVOS instructions, to perform the RefVOS task. Finally, we enhance segmentation robustness through Bidirectional Anchor-updated Propagation (BAP).}
  \label{fig:inference_framework}
  \vspace{-5mm}
\end{figure}

\noindent \textbf{Inference Pipeline:} 
As shown in Fig.~\ref{fig:inference_framework}, inference begins with densely sampled low-resolution frames $I^l_{1:L}$ to compute the frame-query similarity distribution relative to the description. Based on this distribution, the Moment-Centric Sampling (MCS) strategy selects key frames $I^h_{1:N}$. These key frames, together with RefVOS instructions, are then input to the model for segmentation. Finally, Bidirectional Anchor-updated Propagation (BAP) is applied to enhance segmentation robustness.

\subsubsection{Moment-Centric Sampling (MCS)}
\label{subsubsec:mcs}

\begin{figure}[t]
\centering
\vspace{-5mm}
\resizebox{0.9\linewidth}{!}{%
\begin{minipage}{\linewidth}
\begin{algorithm}[H]
\caption{Moment-Centric Sampling (MCS)}
\label{alg:MCS}
\begin{algorithmic}[1]
\State \textbf{Input:} Similarity distribution $\mathcal{S} \in \mathbb{R}^T$, total samples $K$, moment center $c^*$
\State \textbf{Output:} Sampled indices $\mathcal{I} \subset \{0, 1, \ldots, T-1\}$
\State Compute regional weights: $w_L = \sum_{i=0}^{c^*-1} \mathcal{S}_i$, $w_R = \sum_{i=c^*+1}^{T-1} \mathcal{S}_i$
\State Allocate samples proportionally: $k_L = \lfloor (K-1) \cdot \tfrac{w_L}{w_L + w_R} \rceil$, $k_R = K - 1 - k_L$
\State Sample left indices: $\mathcal{I}_L \leftarrow \text{InverseCDF}(\mathcal{S}_{0:c^*-1}, k_L)$
\State Sample right indices: $\mathcal{I}_R \leftarrow \text{InverseCDF}(\mathcal{S}_{c^*+1:T-1}, k_R)$
\State \Return $\text{sort}(\mathcal{I}_L \cup \{c^*\} \cup \mathcal{I}_R)$
\end{algorithmic}
\end{algorithm}
\end{minipage}
}
\vspace{-5mm}
\end{figure}

MCS selectively samples frames most relevant to the text query while preserving global temporal context. It operates by analyzing the similarity distribution between video frames and the query, identifying high-relevance regions, and performing dense sampling around these critical moments while maintaining sparse coverage elsewhere.
Specifically, MCS first computes the similarity between each frame token $I^l_{1:L}$ and the \texttt{[FIND]} token $T^f$. To suppress noise, Gaussian smoothing is applied, producing a refined similarity distribution $\mathcal{S} \in \mathbb{R}^T$. The moment center is then identified via a sliding window that maximizes cumulative similarity within a window of size $w$:
\begin{equation}\label{eqn:window_filter}
\setlength{\abovedisplayskip}{2pt}
\setlength{\belowdisplayskip}{2pt}
i^* = \argmax_{i=0\rightarrow{T-w}} \sum_{j=i}^{i+w-1} \mathcal{S}_j, \quad
c^* = i^* + \lfloor \frac{w}{2} \rfloor,
\end{equation}
where $i^*$ denotes the starting index of the optimal window and $c^*$ the corresponding moment center. This formulation ensures that the most semantically relevant temporal segment is captured within the sequence.
Given the identified moment center $c^*$ and the total number of required samples $K$, moment-centric sampling is performed as outlined in Algorithm~\ref{alg:MCS}. The algorithm partitions the sequence into left and right regions relative to the center, allocates samples proportionally according to cumulative weights, and applies inverse CDF sampling to preserve the similarity distribution. 
The \textit{InverseCDF} procedure takes as input a sub-distribution $\mathcal{S}_{a:b}$ and a target sample size $k$, normalizes the weights to form probabilities $p_i = \mathcal{S}_i / {\sum_{j=a}^b \mathcal{S}_j}$, and computes the $F(i) = \sum_{j=a}^i p_j$. Then, for each uniform random variable $u_m \sim \mathcal{U}(0,1)$, the sampled index is given by:
\begin{equation}
i_m = \min\{\, i \in [a,b] \mid F(i) \ge u_m \,\}, \qquad m=1,\ldots,k.
\end{equation}
This ensures that moments with higher similarity receive denser sampling while still maintaining representation across the entire temporal span. The key advantage of MCS lies in its ability to enhance the capacity to capture both fine-grained actions and their broader temporal dependencies.

\subsubsection{Bidirectional Anchor-updated Propagation (BAP)}
\label{subsubsec:bap}

To improve segmentation robustness, we propose BAP, which leverages key moments as anchors for bidirectional propagation. The primary goal is to provide SAM2 with a strong initialization while ensuring stable tracking through mask updates during propagation. Its two key features are: (1) SAM2 propagation starts from critical temporal anchors and proceeds bidirectionally (\textit{Initialization} in Fig.~\ref{fig:inference_framework}); and (2) mask updates occur at sampled key moments with selective memory clearing (\textit{Update} in Fig.~\ref{fig:inference_framework}). This design enables dynamic error correction, enhancing long-term tracking.
Specifically, the most relevant moment $c^*$ identified by MCS is selected as the starting point for bidirectional propagation. At each sampled key moment $\mathcal{I}$, a memory cleaning mechanism adaptively clears the cache based on the mask prediction score $S^p$ and the tracking score $S^t$, as determined by:
\begin{equation}\label{eqn:amc}
\setlength{\abovedisplayskip}{2pt}
\setlength{\belowdisplayskip}{2pt}
U_i = \begin{cases}
1, & \text{if } \prod_{j=1}^{t} S^t_j < \lambda \cdot S^p_i \\
0, & \text{otherwise}
\end{cases}
\end{equation}
where $U_i$ is the binary indicator (1 for clearing memory, 0 for retention), $S^t_j$ is the tracking score at step $j$, $\prod_{j=1}^{t} S^t_j$ denotes the cumulative tracking confidence up to time $t$, $S^p_i$ is the prediction confidence at node $i$, and $\lambda$ is a sensitivity hyperparameter (set to 0.9). An update is triggered when accumulated confidence falls below the threshold determined by the current prediction quality, ensuring that memory is refreshed only when beneficial for tracking accuracy.

\section{Experiments}
\label{sec:experiment}

\subsection{Implementation Details}
\label{subsec:implementation_details}
We construct our baseline by integrating Qwen2.5-VL~\citep{Qwen2.5-VL} with SAM2~\citep{ravi2024sam2}. Following prior work~\citep{yuan2025sa2va}, segmentation masks are generated by decoding the hidden state of the \texttt{[SEG]} token using the SAM2 decoder. The LLM is fine-tuned with LoRA~\citep{hu2021lora}, with a maximum sequence length of 8,192 tokens. Training is conducted on 8 NVIDIA H20 GPUs (96GB memory each) and takes approximately 24 hours for the MomentSeg-3B model. For evaluation, we adopt lmms-eval~\citep{zhang2024lmmsevalrealitycheckevaluation} for both image and video chat tasks.

\subsection{Main Results}
\noindent \textbf{Temporal Sentence Grounding Task.}
We evaluate our proposed method, MomentSeg, on the Charades-STA and ActivityNet datasets. As shown in Table~\ref{tab:results_tsg}, MomentSeg is compared with a range of advanced methods. The MomentSeg-3B model, despite its smaller scale, consistently outperforms most existing Video-LLMs. This strong performance validates the effectiveness of our proposed paradigm, which is underpinned by the novel mechanism where the \texttt{[FIND]} token interacts with temporal tokens via cosine similarity to achieve frame-level matching.

\begin{table}[t]
    \centering
    \vspace{-2mm}
    \caption{Comparison of MomentSeg with methods on temporal sentence grounding benchmarks.}
    \vspace{-2mm}
    \resizebox{1.0\linewidth}{!}{
        \begin{tabular}{l|ccc|c|ccc|c}
        \toprule
        \multirow{2}{*}{Method} &\multicolumn{4}{|c}{Charades-STA} &\multicolumn{4}{|c}{ActivityNet-Grounding}\\
        \cmidrule(lr){2-5} \cmidrule(lr){6-9}
        &R@0.3 &R@0.5 &R@0.7 &mIoU &R@0.3 &R@0.5 &R@0.7 &mIoU \\
        \midrule
        Video-LLaMA-7B \citep{zhang2023video}               & 25.2 & 10.6 & 3.4  & {16.8} & 21.9 & 10.8 & 4.9  & {16.5} \\
        SeViLA-3B \citep{yu2023self}                        & 27.0 & 15.0 & 5.8  & \cellcolor{gray!10}{18.3} & 31.6 & 19.0 & 10.1 & \cellcolor{gray!10}{23.0} \\
        Video-ChatGPT-7B \citep{maaz2023video}              & 27.2 & 6.2  & 1.9  & {19.7} & 19.5 & 10.6 & 4.8  & {14.2}\\
        VideoChat2-7B \citep{li2024mvbench}                  & 38.0 & 14.3 & 3.8  & \cellcolor{gray!10}{24.6} & 40.8 & 27.8 & 9.3  & \cellcolor{gray!10}{27.9} \\
        Momenter-7B \citep{qian2024momentor}                 & 42.6 & 26.6 & 11.6 & \cellcolor{gray!10}{28.5} & 42.9 & 23.0 & 12.4 & \cellcolor{gray!10}{29.3} \\
        VTimeLLM-7B \citep{huang2024vtimellm}                & 51.0 & 27.5 & 11.4 & \cellcolor{gray!10}{31.2} & 44.0 & 27.8 & {14.3} & \cellcolor{gray!10}{30.4} \\
        TimeChat-7B \citep{ren2024timechat}                  & -    & 32.2 & 13.4 & \cellcolor{gray!10}-    & -    & -    & -         & \cellcolor{gray!10}-   \\
        VTG-LLM-7B \citep{guo2025vtg}                        & -    & 33.8 & 15.7 & \cellcolor{gray!10}-    & -    & -    & -           & \cellcolor{gray!10}-    \\
        HawkEye-7B \citep{wang2024hawkeye}                   & 50.6 & 31.4 & 14.5 & \cellcolor{gray!10}{33.7} & 49.1 & 29.3 & 10.7 & \cellcolor{gray!10}{32.7}\\
        Grounded-VideoLLM-4B \citep{wang2024grounded}        & 54.2 & 36.4 & 19.7 & \cellcolor{gray!10}{36.8} & 46.2 & 30.3 & 19.0 &\cellcolor{gray!10}{36.1} \\  
        NumPro-LongVA-7B~\citep{numberit}                    & \underline{63.8} & \underline{42.0} & \underline{20.6} & \cellcolor{gray!10}{41.4} & \underline{55.6} & \underline{37.5} & \underline{20.6} & \cellcolor{gray!10}{\underline{38.8}} \\
        \midrule 
        Qwen2.5-VL-7B \citep{Qwen2.5-VL}                      & - & -  &  - & \cellcolor{gray!10}{\underline{43.6}} & - & -  & - & \cellcolor{gray!10}- \\
        \midrule
        \textbf{MomentSeg-3B (Ours)}                         & \textbf{76.1} & \textbf{58.2} & \textbf{25.8} & \cellcolor{gray!10}{\textbf{50.2}} & \textbf{67.5} & \textbf{44.7} & \textbf{23.2} & \cellcolor{gray!10}{\textbf{45.4}} \\
        \bottomrule 
    \end{tabular}
    }
    \vspace{-2mm}
    \label{tab:results_tsg}
\end{table}

\begin{table}[t]
	\centering
  %  \vspace{-2mm}
    \caption{Comparison of MomentSeg with methods on referring video object segmentation benchmarks.}
    \vspace{-2mm}
	\resizebox{1.0\textwidth}{!}{
		\begin{tabular}{l|ccc|ccc|ccc|ccc}
			\toprule
			\multirow{2}{*}{Method}  & \multicolumn{3}{c|}{MeViS ($val^u$)} & \multicolumn{3}{c|}{MeViS ($val$)} &\multicolumn{3}{c|}{Ref-Youtube-VOS} & \multicolumn{3}{c}{Ref-DAVIS17}  \\
            \cmidrule(lr){2-4} \cmidrule(lr){5-7} \cmidrule(lr){8-10} \cmidrule(lr){11-13}
			& $\mathcal{J}$\&$\mathcal{F}$ & $\mathcal{J}$ & $\mathcal{F}$& $\mathcal{J}$\&$\mathcal{F}$& $\mathcal{J}$ & $\mathcal{F}$& $\mathcal{J}$\&$\mathcal{F}$ & $\mathcal{J}$ & $\mathcal{F}$ & $\mathcal{J}$\&$\mathcal{F}$ & $\mathcal{J}$ & $\mathcal{F}$  \\ 
			% \midrule
      %       \multicolumn{13}{c}{\emph{Methods without LLMs}} \\
      %       \midrule
      %       % URVOS~\citep{seo2020urvos}  &&& &27.8 & 25.7 & 29.9 & 47.2 & 45.2 & 49.1   \\
			%    ReferFormer~\citep{wu2022referformer}  &-&-&- & 31.0 &29.8 &32.2 & 59.4 & 58.0 & 60.9  & 59.6 & 56.5 & 62.7\\
      %       LMPM~\citep{ding2023mevis}  & 40.2 & 36.5 & 43.9 & 37.2 & 34.2 & 40.2 & -& -& -  & -& -& -       \\
      %       DsHmp~\citep{DsHmp}  &55.3 &51.0 & 60.4 & 46.4 & 43.0 & 49.8 & 63.6 & 61.8 &65.4 & 64.0 & 60.8 & 67.2    \\
      %       DMVS~\citep{ding2023mevis} & 58.3 & 52.6 & 63.9 & 48.6 & 44.2 & 52.9 & 64.3 & 62.4 & 66.2 & 65.2 & 62.2 & 68.2\\
      %       \midrule
      %       % \rowcolor{tablegray}
      %       \multicolumn{13}{c}{\emph{Methods with LLMs}} \\
            \midrule
			      LISA-7B~\citep{lai2024lisa}                       & \cellcolor{gray!10}{43.2} & 39.9 & 46.5 & \cellcolor{gray!10}{37.2}& 35.1& 39.4   & \cellcolor{gray!10}{53.9}& 53.4&54.3    & \cellcolor{gray!10}{64.8} & 62.2 & 67.3 \\
			      % LISA-13B~\citep{lai2024lisa}                    &37.9&35.8&40.0&54.4&54.0&54.8\\
			      TrackGPT-7B~\citep{zhu2023tracking}               & \cellcolor{gray!10}- & - & -            &\cellcolor{gray!10}{40.1} &37.6 &42.6    &\cellcolor{gray!10}{56.4}  &55.3  &57.4  & \cellcolor{gray!10}{63.2} & 59.4 & 67.0 \\
			      % TrackGPT-13B~\citep{zhu2023tracking}            &41.2  &39.2  &43.1 & 59.5 & 58.1 & 60.8\\
            % VideoGLaMM-7B~\citep{munasinghe2024videoglamm}  & - &- &-  & 45.2 & 42.1 & 48.2 & 66.8 & 65.4 & 68.2  & 69.5 & 73.3 & 65.6  \\
            VideoLISA-3.8B~\citep{bai2024one}                 & \cellcolor{gray!10}{54.5} & 50.9 & 58.1 & \cellcolor{gray!10}{44.4} & 41.3 & 47.6 & \cellcolor{gray!10}{63.7} & 61.7 & 65.7 & \cellcolor{gray!10}{68.8} & 64.9 & 72.7\\
            VISA-7B~\citep{yan2024visa}                       & \cellcolor{gray!10}- & - & -            & \cellcolor{gray!10}{43.5} & 40.7 & 46.3 & \cellcolor{gray!10}{61.5} & 59.8 & 63.2 & \cellcolor{gray!10}{69.4} & 66.3 & 72.5 \\
            % VISA-13B~\citep{yan2024visa}                    & 44.5 & 41.8 & 47.1 & 63.0 & 61.4 & 64.7\\
            ViLLa-6B~\citep{zheng2024villa}                   & \cellcolor{gray!10}- & - & -            & \cellcolor{gray!10}{49.4} &46.5 &52.3   & \cellcolor{gray!10}{67.5} &64.6 &70.4   &\cellcolor{gray!10}{74.3}& 70.6 &78.0\\
            {GLUS}-7B~\citep{lin2025glus}                     & \cellcolor{gray!10}{60.9} & - & -       & \cellcolor{gray!10}{51.3} &48.5&54.2    & \cellcolor{gray!10}{67.3} &65.5 &69.0   & \cellcolor{gray!10}- & - & - \\
            InstructSeg-3B~\citep{wei2024instructseg} & \cellcolor{gray!10}- & - & - & \cellcolor{gray!10}- & - & - & \cellcolor{gray!10}67.5 & 65.4 & 69.5 & \cellcolor{gray!10}71.1 & 67.3 & 74.9 \\
            Sa2VA-4B~\citep{yuan2025sa2va}                    & \cellcolor{gray!10}- & - & -            & \cellcolor{gray!10}{46.2} & - & -       & \cellcolor{gray!10}{70.0} & - & -       & \cellcolor{gray!10}{70.0} & - & - \\
            Sa2VA-8B~\citep{yuan2025sa2va}                    & \cellcolor{gray!10}- & - & -            & \cellcolor{gray!10}{46.9} & - & -       & \cellcolor{gray!10}{70.7} & -&-         & \cellcolor{gray!10}{75.2} & - & -\\
            \midrule
            Sa2VA-Qwen2.5-VL-3B          & \cellcolor{gray!10}57.5 & 53.6 & 61.3   & \cellcolor{gray!10}50.0 &  46.9& 53.1            & \cellcolor{gray!10}71.0 & 68.8&73.1              & \cellcolor{gray!10}74.4& 70.1 & 78.7\\ % 5frames used in sa2va
            \midrule
            \textbf{MomentSeg-3B (Ours)}                      & \cellcolor{gray!10}{\underline{62.0}} & \underline{58.1} & \underline{65.9} & \cellcolor{gray!10}{\underline{54.8}} & \underline{51.7} & \underline{58.0} & \cellcolor{gray!10}{\underline{72.0}} & \underline{69.8} & \underline{74.3} & \cellcolor{gray!10}{\underline{76.4}} & \underline{72.2} & \underline{80.6} \\
            \textbf{MomentSeg-7B (Ours)}                      & \cellcolor{gray!10}{\textbf{62.6}} & \textbf{58.7} & \textbf{66.5} & \cellcolor{gray!10}{\textbf{57.1}} & \textbf{53.9} & \textbf{60.2} & \cellcolor{gray!10}{\textbf{72.3}} & \textbf{70.1} & \textbf{74.5} & \cellcolor{gray!10}{\textbf{77.4}} & \textbf{73.2} & \textbf{81.7}\\
            % \textbf{MomentSeg-32B (Ours)} & \\
            \bottomrule
	\end{tabular} 
        }
        \vspace{-3mm}
	\label{tab:refvos_main_results}
\end{table}

% without R
\begin{table}[t]
	\centering
   \vspace{-4mm}
    \caption{Comparison of MomentSeg with methods on reasoning video object segmentation benchmarks.}
    \vspace{-2mm}
	\resizebox{0.95\textwidth}{!}{
		\begin{tabular}{l|ccc|ccc|ccc|ccc}
			\toprule
			\multirow{3}{*}{Method}  & \multicolumn{9}{c|}{ReVOS} & \multicolumn{3}{c}{ReasonVOS} \\ 
            \cmidrule(lr){2-10} \cmidrule(lr){11-13}
            & \multicolumn{3}{c|}{Referring} &\multicolumn{3}{c|}{Reasoning} &\multicolumn{3}{c|}{Overall}  &  \multirow{2}{*}{$\mathcal{J}$\&$\mathcal{F}$}& \multirow{2}{*}{$\mathcal{J}$} & \multirow{2}{*}{$\mathcal{F}$}\\ 

			& $\mathcal{J}$\&$\mathcal{F}$ & $\mathcal{J}$ & $\mathcal{F}$ & $\mathcal{J}$\&$\mathcal{F}$ & $\mathcal{J} $ &$\mathcal{F}$ & $\mathcal{J}$\&$\mathcal{F}$ & $\mathcal{J}$ & $\mathcal{F}$ & & & \\
            % \midrule
            %     \multicolumn{14}{c}{\emph{Methods without LLMs}} \\
            % \midrule
            % % MTTR~\citep{tang2023temporal} & 29.8 &30.2 &30.0 &20.4 &21.5 &21.0 &25.1 &25.9 &25.5 & 5.6& 31.1 & 29.1 & 33.1\\
            % ReferFormer~\citep{wu2022referformer} & 31.2 & 34.3 & 32.7 & 21.3 & 25.6 & 23.4 & 26.2 & 29.9 & 28.1 & 8.8 & 32.9 & 30.2 & 35.6 \\
            % SOC~\citep{luo2023soc} & - & - & - & - & - & - & - & - & - & - & 35.9 & 33.3 & 38.5\\
            % \midrule
            %     \multicolumn{14}{c}{\emph{Methods with LLMs}} \\
            \midrule
            % LISA-13B~\citep{lai2024lisa} & \cellcolor{gray!10}46.6 & 45.2    & 47.9 & \cellcolor{gray!10}36.7 & 34.3   & 39.1 & \cellcolor{gray!10}41.6 & 39.8   & 43.5  & \cellcolor{gray!10}- & - & - \\
            TrackGPT-13B~\citep{zhu2023tracking} &\cellcolor{gray!10}49.5 &48.3&50.6& \cellcolor{gray!10}40.5 & 38.1  & 42.9  & \cellcolor{gray!10}45.0 & 43.2  & 46.8   & \cellcolor{gray!10}- & - & - \\
            VISA-13B~\citep{yan2024visa}  & \cellcolor{gray!10}57.4 & 55.6 & 59.1   & \cellcolor{gray!10}44.3 & 42.0 & 46.7   & \cellcolor{gray!10}50.9 & 48.8 & 52.9    & \cellcolor{gray!10}- & - & - \\
            VideoLISA-3.8B~\citep{bai2024one} & \cellcolor{gray!10}- & - & -             & \cellcolor{gray!10}- & - & -            & \cellcolor{gray!10}- & - & -             & \cellcolor{gray!10}47.5 & 45.1 & 49.9 \\
            HyperSeg-3B~\citep{wei2024hyperseg} &\cellcolor{gray!10}58.5&56.0& 60.9 & \cellcolor{gray!10}53.0 & 50.2   & 55.8 & \cellcolor{gray!10}55.7 & 53.1   & 58.4  & \cellcolor{gray!10}- & - & - \\
            ViLLa-6B~\citep{zheng2024villa} & \cellcolor{gray!10}- & - & -          & \cellcolor{gray!10}- & - & -            & \cellcolor{gray!10}57.0 & 54.9 & 59.1    & \cellcolor{gray!10}55.4& - & - \\
            GLUS-7B~\citep{lin2025glus} & \cellcolor{gray!10}58.3 & 56.0 & 60.7     & \cellcolor{gray!10}51.4 & 48.8 & 53.9   & \cellcolor{gray!10}54.9 & 52.4 & 57.3    & \cellcolor{gray!10}{49.9} & {47.5} & {52.4} \\
            Sa2VA-4B~\citep{yuan2025sa2va} & \cellcolor{gray!10}- & - & -           & \cellcolor{gray!10}- & - &-             & \cellcolor{gray!10}53.2 & - & -          & \cellcolor{gray!10}- & - & - \\
            VRS-HQ-7B~\cite{gong2025devil} & \cellcolor{gray!10}62.1 & 59.8 & 64.5  & \cellcolor{gray!10}56.1 & 53.5 & 58.7   & \cellcolor{gray!10}59.1 & 56.6 & 61.6    & \cellcolor{gray!10}- & - & - \\
            \midrule
            Sa2VA-Qwen2.5-VL-3B& \cellcolor{gray!10}61.5 & 59.0 & 64.1  & \cellcolor{gray!10}55.9 & 53.0 & 58.8  & \cellcolor{gray!10}58.7 & 56.0&61.4  & \cellcolor{gray!10}49.6& 46.9& 52.4\\ % 5frames used in sa2va
            \midrule
            \textbf{MomentSeg-3B (Ours)} & \cellcolor{gray!10}\underline{65.4} & \underline{63.0} & \underline{67.9} & \cellcolor{gray!10}\underline{59.9} & \underline{57.1} & \underline{62.6} & \cellcolor{gray!10}\underline{62.6} & \underline{60.0} & \underline{65.2}         & \cellcolor{gray!10}\underline{61.7} & \underline{58.2} & \underline{65.3} \\
            \textbf{MomentSeg-7B (Ours)} & \cellcolor{gray!10}\textbf{67.4} & \textbf{64.9} & \textbf{69.8} & \cellcolor{gray!10}\textbf{62.8} & \textbf{59.8} & \textbf{65.7} & \cellcolor{gray!10}\textbf{65.1} & \textbf{62.3} & \textbf{67.8} & \cellcolor{gray!10}\textbf{62.7} & \textbf{59.2} & \textbf{66.1}\\
            % \textbf{MomentSeg-32B (Ours)} & 67.1 & 64.6 & 69.6 & 63.2 & 60.4 &  65.9 & 65.2 & 62.5 & 67.8 & 91.4& 62.6 & 58.9 & 66.2 \\
            \bottomrule
	\end{tabular} 
        }
        \vspace{-2mm}
	\label{tab:reasonvos_main_results}
\end{table}

\textbf{Referring/Reasoning Video Segmentation Task.}
For \textit{referring} tasks (Table~\ref{tab:refvos_main_results}), MomentSeg-3B achieves $\mathcal{J}$\&$\mathcal{F}$ scores of 62.0 on MeViS ($val^u$) and 54.8 on MeViS ($val$), surpassing GLUS-7B by 1.1 and 3.5 points, respectively. On Ref-Youtube-VOS and Ref-DAVIS17, it attains scores of 72.0 and 76.4, improving over Sa2VA-4B by 2.0 and 6.4 points. Scaling to 7B further enhances performance across benchmarks.  
For \textit{reasoning} tasks (Table~\ref{tab:reasonvos_main_results}), MomentSeg-3B obtains 62.6 on ReVOS, exceeding VRS-HQ-7B by 3.5 points. On ReasonVOS, it achieves 61.7, outperforming ViLLa-6B by 6.3 points, with consistent improvements at larger scales.

\noindent \textbf{Referring/Reasoning Image Segmentation Task.}  
As shown in Table~\ref{tab:image_segmentation}, MomentSeg-3B achieves 82.1, 76.9, and 78.8 on the RefCOCO/+/g validation sets, outperforming Sa2VA-4B by 3.2, 5.2, and 4.7 points, respectively. For reasoning tasks, MomentSeg-7B attains 65.5 on the ReasonSeg test set, surpassing the previous method, LISA-7B, by 2.6 points. Furthermore, on the GCG benchmark, MomentSeg-7B achieves 67.8/67.9 mIoU (val/test), exceeding GLaMM-7B by 2.0/3.3 points.

\begin{table}[t]
	\centering
   \vspace{-2mm}
	\caption{Comparison of model results across image segmentation tasks. Both RefCOCO/+/g and ReasonSeg use the cIoU metric, while GCG employs the mIoU metric.}
    \vspace{-2mm}
    \resizebox{1.0\textwidth}{!}{ 
	\begin{tabular}{l|ccc|ccc|cc|cc|cc}
		\toprule
		% Method & RefCOCO & RefCOCO+ & RefCOCOg & GCG & ReasonSeg \\
        \multirow{2}{*}{Method}  & \multicolumn{3}{c|}{RefCOCO} & \multicolumn{3}{c|}{RefCOCO+} & \multicolumn{2}{c|}{RefCOCOg} & \multicolumn{2}{c|}{ReasonSeg} & \multicolumn{2}{c}{GCG} \\
        % \cline{2-4}
        \cmidrule(lr){2-4} \cmidrule(lr){5-7} \cmidrule(lr){8-9} \cmidrule(lr){10-11} \cmidrule(lr){12-13}
        & val & testA & testB & val & testA & testB & val(U) & test(U) & val & test & val & test \\
        \midrule
      %       \multicolumn{11}{l}{\emph{Methods without LLMs}} \\
      %   \midrule
      %   ReLA~\citep{liu2023gres}  & 73.8 & 76.5 & 70.2  & 66.0 &71.0 &57.7& 65.0& 66.0& - & - & - & - \\
      %   DeRIS~\citep{dai2025deris} & 77.8 & 79.1 & 76.0 & 72.2 & 75.4 & 67.9 & 72.2 & 73.0 & - & - & - & -  \\
        % SEEM~\citep{zou2023segment} & - & -  & - & - & - & -  & 65.7 & - & 25.5 & 21.2& - & - \\
      %   \midrule
      %       \multicolumn{11}{l}{\emph{Methods with LLMs}} \\
      %   \midrule
        PixelLM-7B~\citep{ren2024pixellm} & 73.0 & 76.5 & 68.2 & 66.3 & 71.7 & 58.3 & 69.3 & 70.5 & - & - &-&-\\
        % LaSagnA~\citep{wei2024lasagna} & 76.8 & 66.4 & 70.6 & & \\
		   LISA-7B~\citep{lai2024lisa} & 74.9 & 79.1 & 72.3 & 65.1 & 70.8 & 58.1 & 67.9 & 70.6 & 61.3 & 62.9 & 62.0 & 61.7 \\
		   GLaMM-7B~\citep{hanoona2023GLaMM} & 79.5 & 83.2 & 76.9 & 72.6 & 78.7 & 64.6 & 74.2 & 74.9 & & & 65.8 & 64.6 \\
        VisionLLMv2-7B~\citep{wu2024visionllm} & 79.2 & 82.3 & 77.0 & 68.9 & 75.8 & 61.8 & 73.3 & 74.8 & 56.9 & 48.3 & 64.6 & 62.8 \\
        OMG-LLaVA-7B~\citep{zhang2024omg} &75.6 & 77.7 & 71.2 & 65.6 & 69.7 & 58.9 & 70.7 & 70.2 & -& -&65.5&64.7 \\
        LaSagnA-7B~\citep{wei2024lasagna} & 76.8 & 78.7 & 73.8 & 66.4 & 70.6 & 60.1 & 70.6 & 71.9 & 48.8 & 47.2 & - & -\\
        VISA-7B~\citep{yan2024visa} & 72.4 & 75.5 & 68.1 & 59.8 & 64.8 & 53.1 & 65.5 & 66.4 & 52.7 & 57.8 & -& -\\
        VRS-HQ-7B~\citep{gong2025devil} & 73.5 & 77.5 & 69.5 & 61.7 & 67.6 & 54.3 & 66.7 & 67.5 & 51.8 & 52.9 & - & - \\
  
		   Sa2VA-4B~\citep{yuan2025sa2va} & 78.9 & - & - & 71.7 & - & - & 74.1 & - & - & - & -&- \\
      %   Sa2VA-8B~\citep{yuan2025sa2va} & 81.6 & - & - & 76.2& - & - & 78.7& - & - & - & - & - \\
        \midrule
        \textbf{MomentSeg-3B (Ours)} & \underline{82.1} & \underline{83.7} & \underline{79.2} & \underline{76.9} & \underline{81.1} & \underline{71.8} & \underline{78.8} & \underline{79.2} & \underline{62.0} & \underline{64.3} & \underline{67.0} & \underline{65.9} \\
        \textbf{MomentSeg-7B (Ours)} & \textbf{82.6} & \textbf{85.1} & \textbf{80.2} & \textbf{78.2} & \textbf{81.9} & \textbf{72.3} & \textbf{80.1} & \textbf{80.1} & \textbf{63.3} & \textbf{65.5} & \textbf{67.8} & \textbf{67.9}\\ 
        % \textbf{MomentSeg-32B (Ours)} & 82.9 & 84.8 & 80.5 & 78.1 & 81.9 & 73.2 & 80.6 & 80.4 & 64.4 & 68.1 & \\
		\bottomrule
	\end{tabular}
    }
	\label{tab:image_segmentation}
   \vspace{-4mm}
\end{table}

\subsection{Ablations Studies}
\label{sec:ablation}

\textbf{Key Components Design.}
To assess the effectiveness of individual components, we conduct ablation studies on six RefVOS datasets. As shown in Table~\ref{table:ablation_components}, starting from the baseline, introducing Temporal Token Injecting (TTI) consistently improves performance, with smaller gains on Ref-Youtube-VOS and Ref-DaVIS17 due to shorter videos and static expression. Incorporating MCS provides further improvements, particularly on motion-intensive datasets such as MeViS, where key frame sampling enhances action understanding. Finally, adding BAP strengthens tracking and error correction, with the largest benefits observed on reasoning-oriented benchmarks involving long video sequences, underscoring the importance of robust temporal tracking mechanisms.

\textbf{Effectiveness of Moment-Centric Sampling.}
Different sampling strategies affect performance in dataset-specific ways. MeViS requires action understanding in long videos where targets appear only in certain segments, Ref-DAVIS17 involves targets that are present throughout most of the video, and ReasonVOS emphasizes relational reasoning with targets often emerging mid-video. As shown in Table~\ref{table:ab_sampler}, \textit{FirstK} underperforms relative to \textit{Uniform} on MeViS and ReasonVOS because early frames frequently miss target segments. \textit{Keyframe}–based strategies alleviate this by prioritizing high-similarity moments: \textit{NearbyK} samples adjacent frames, while \textit{TopK} selects globally most similar frames. Among them, \textit{NearbyK} proves more robust, underscoring the benefit of dense sampling around critical regions. 
Building on this insight, our proposed \textit{Moment-Centric Sampling} integrates dense sampling near key moments with sparse sampling elsewhere, thereby capturing both critical and global context and achieving the best overall performance across benchmarks.

\textbf{Effectiveness of Bidirectional Anchor-updated Propagation.}  
Mask propagation often struggles with poor initialization when the target appears late in the video. BAP overcomes this by initializing at the most probable target moment, propagating bidirectionally, and applying adaptive memory updates to limit error accumulation.  
Table~\ref{table:ab_bap} shows ablations of its components. \textit{Moment-anchored Updating} notably improves performance, particularly on long video scenarios, by correcting accumulated tracking errors, while \textit{Adaptive Memory Cleaning} provides smaller but consistent improvements. Overall, \textit{Bidirectional Propagation} strategy improves robustness by initializing from the most probable target moment without adding inference cost.

\textbf{Effectiveness of Temporal Tokens for TSG and RefVOS.}  
We use temporal tokens in training for both tasks, with RefVOS additionally sampling frames for segmentation. We examine: \textbf{(1)} whether TTI in RefVOS establishes temporal action cues; \textbf{(2)} whether tokens can be omitted at inference to reduce overhead; and \textbf{(3)} whether jointly optimizing temporal tokens for RefVOS and TSG leads to mutual benefits.
For (1) and (2), Table~\ref{tab:ab_video_tokens} shows consistent gains across datasets, including +2 points on ReasonVOS. Notably, inference without tokens achieves similar performance, so we omit them to avoid extra cost. More detailed analyses are in Appendix~\ref{app_subsubsec:ablation_temporal_tokens}.  
For (3), Table~\ref{tab:ab_joint_training} demonstrates that joint training improves both tasks, yielding a 1.2-point gain on TSG.  
Jointly optimizing temporal tokens provides mutual benefits, and incorporating the video-chat data further boosts performance.

\begin{table}[t]
    \centering
    \caption{Effectiveness of the key components. 
   %  TTJ denotes Temporal Tokens Injecting, MCS represents Moment-Centric Sampling, and BAP stands for Bidirectional Action-anchored Propagation. 
    We report the $\mathcal{J}$\&$\mathcal{F}$ metric for all six datasets.}
    \vspace{-2mm}
    \resizebox{0.85\textwidth}{!}{
    \begin{tabular}{c|c|c|c|c|c|c|c|c|c}
        \toprule
        % TTI: Temporal Tokens Injecting, KCAS: Key-Context Adaptive Sampling, BAP: Bidirectional Action-anchored Propagation
        {ID} &{TTI} & {MCS}& {BAP} & MeViS ($val^u$) & {MeViS ($val$)} & {Ref-DAVIS17} & {Ref-Youtube-VOS} & {ReasonVOS} & {ReVOS (Overall)}\\
        % &  &  &  & $\mathcal{J}$\&$\mathcal{F}$ & $\mathcal{J}$\&$\mathcal{F}$ & $\mathcal{F}$  & $\mathcal{J}$\&$\mathcal{F}$ & $\mathcal{J}$ & $\mathcal{F}$ \\
        \midrule
        1 &             &              &              &  57.0 & 51.3 & 75.5  & 70.1 & 54.1 & 58.3 \\
        2 & \checkmark  &              &              &  58.7(\up{1.7}) & 52.3(\up{1.0}) & 76.2(\up{0.7}) & 70.3(\up{0.2}) & 58.1(\up{4.0})  & 59.2(\up{0.9}) \\
        3 & \checkmark  & \checkmark   &              &  61.3(\up{2.6}) & 53.9(\up{1.6}) & 76.8(\up{0.6})  & 70.8(\up{0.5}) & 60.8(\up{2.7}) & 60.4(\up{1.2})\\
        4 & \checkmark  & \checkmark   & \checkmark   &  62.0(\up{0.7}) & 54.1(\up{0.2}) & 76.4(\down{0.4})  & 72.0(\up{1.2}) & 61.7(\up{0.9}) & 62.6(\up{2.2}) \\
        \bottomrule
    \end{tabular}
    }
    \label{table:ablation_components}
    \vspace{-2mm}
\end{table}
\begin{table}[t]
  \centering
  \begin{minipage}[t]{0.46\textwidth}
        \centering
        \caption{Impact of moment-centric sampling.}
        \vspace{-2mm}
        \resizebox{\textwidth}{!}{
            \begin{tabular}{l|c|c|c}
                \toprule
                {Method}  & {MeViS ($val^u$)} & {Ref-DAVIS17} & {ReasonVOS}   \\
                \midrule
                FirstK (Baseline)  & 58.3 & 76.3 & 55.9 \\
                Uniform  & 59.6(\up{1.3}) & 76.2(\down{0.1}) & 57.1 (\up{1.2})\\
                \midrule
                KeyFrame (TopK)  & 60.1 & 76.9 & 60.2 \\
                KeyFrame (NearbyK)  & 60.9 (\up{0.8}) & 77.1(\up{0.2}) & 59.9(\down{0.3}) \\
                \midrule
                MCS & 61.3(\up{3.0}) & 77.3(\up{1.0}) & {60.8}(\up{4.9})\\
                \bottomrule
            \end{tabular}
            \vspace{-2mm}
        }
        \vspace{-5mm}
        \label{table:ab_sampler}
      \end{minipage}\hfill
    \begin{minipage}[t]{0.51\textwidth}
      \centering
      \caption{Effectiveness of bidirectional anchor-updated propagation. $\mathcal{J}$\&$\mathcal{F}$ metrics are reported.}
      \vspace{-2mm}
      \resizebox{\textwidth}{!}{
        \begin{tabular}{l|c|c|c}
          \toprule
          {Method}  & {MeViS ($val^u$)} & {Ref-DAVIS17} & {ReasonVOS}   \\
          \midrule
          Forward Propgation (Baseline) & 58.9 & 76.2 & 57.4 \\
          \midrule
          +Moment-anchored Updating  & 61.1(\up{2.2}) & {76.9}(\up{0.7}) & 60.8(\up{3.4}) \\
          +Adaptive Memory Cleaning  & 61.3(\up{0.2}) & 77.0(\up{0.1}) & 61.0(\up{0.2}) \\
          +Bidirectional Propgation & {62.0}(\up{0.7}) & 76.8(\down{0.2}) & {61.7}(\up{0.7}) \\
          \bottomrule
        \end{tabular}
        \vspace{-2mm}
      }
      \label{table:ab_bap}
    \end{minipage}
    \vspace{-5mm}
\end{table}

\begin{table}[H]
    \centering
    \begin{minipage}[t]{0.47\textwidth}
        \caption{Effectiveness of temporal token injecting.}
        \vspace{-2mm}
        \centering
        \resizebox{\textwidth}{!}{
            \begin{tabular}{l|c|c|c}
                \toprule
                {Method}  & {MeViS ($val^u$)} & {Ref-DAVIS17} & {ReasonVOS}   \\
                % & $\mathcal{J}$\&$\mathcal{F}$& $\mathcal{J}$\&$\mathcal{F}$ &  $\mathcal{J}$\&$\mathcal{F}$ \\ 
                \midrule
                \multicolumn{4}{c}{\emph{Training Process}} \\
                w/o Temporal Tokens  & 59.1 & 73.3 & 53.1 \\
                w. Temporal Tokens  & 60.1 (\up{1.0}) & 74.2(\up{0.9}) & 55.1(\up{2.0}) \\
                \midrule
                \multicolumn{4}{c}{\emph{Inference Process}} \\
                w/o Temporal Tokens & 60.1 & 74.2 &  55.1   \\
                w. Temporal Tokens & 60.3 (\up{0.2}) & 73.2(\down{1.0}) & 56.4(\up{1.3}) \\
                \bottomrule
            \end{tabular}
        }
        \label{tab:ab_video_tokens}
    \end{minipage}
    \hfill
    % \vspace{2mm}
    \begin{minipage}[t]{0.52\textwidth}
    \centering
    \caption{Effectiveness of joint training across tasks.}
    \vspace{-2mm}
    \resizebox{\textwidth}{!}{
      \begin{tabular}{ccc|cc|c}
        \toprule
        \multicolumn{3}{c|}{Training Data} & \multicolumn{2}{c|}{RefVOS} & {TSG} \\
        \cmidrule(lr){1-3} \cmidrule(lr){4-5} \cmidrule(lr){6-6}
        RefVOS & TSG & VideoChat & {MeViS} & {ReVOS} & {Charades-STA} \\
        \midrule
        \checkmark  &             &  & 50.7 & 58.9 & -  \\
                    & \checkmark  &  & - & - & 47.7 \\
        \midrule
        \checkmark  & \checkmark  &  & 51.4(\up{0.7}) & 59.8(\up{0.9}) & 48.9(\up{1.2}) \\
        \checkmark  & \checkmark  & \checkmark & {52.7}(\up{1.3}) & {60.3}(\up{0.5}) & {49.3}(\up{0.4}) \\
        \bottomrule
      \end{tabular}
    }
      \label{tab:ab_joint_training}
   \end{minipage}
   \vspace{-4mm}
\end{table}

\section{Qualitative Results}
\label{sec:qualitative_results}
\begin{figure}[h]
\centering
\vspace{-3mm}
\includegraphics[width=0.95\linewidth]{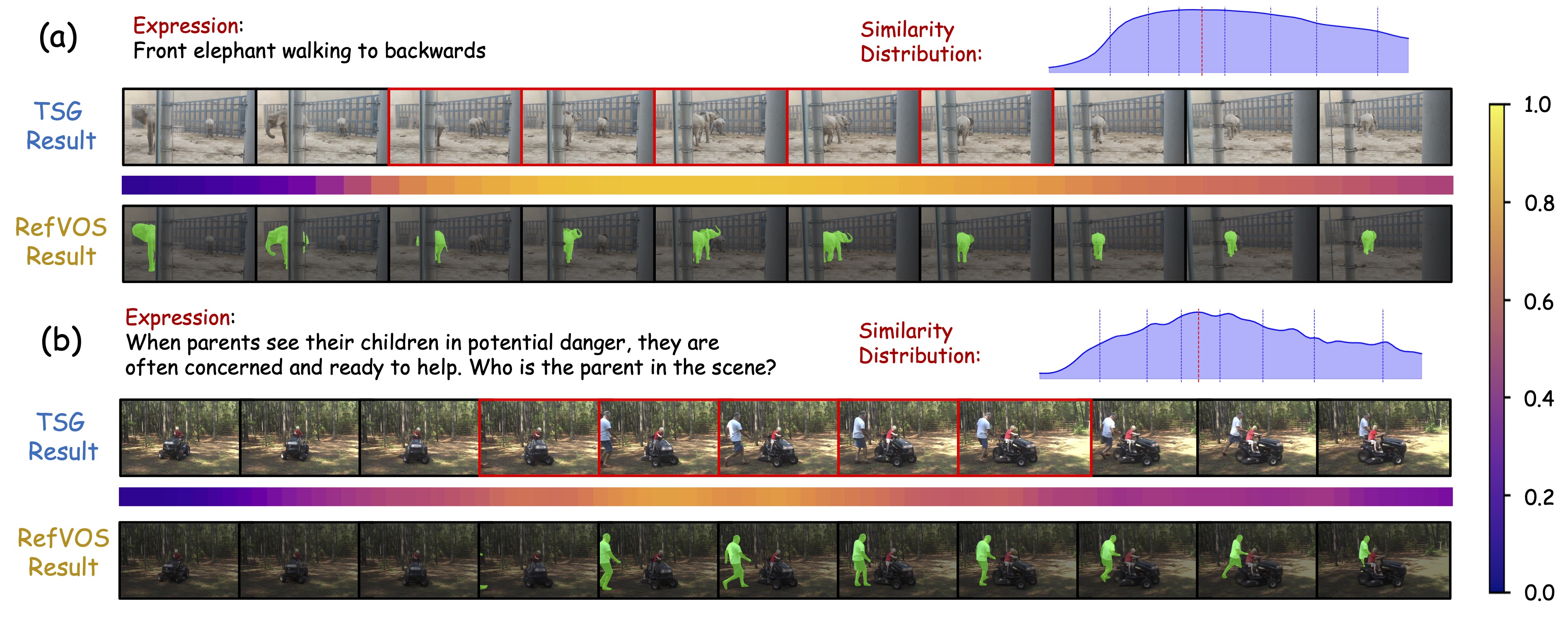}
\vspace{-2mm}
\caption{Qualitative results of MomentSeg on TSG and RefVOS tasks. The figure displays the input expression, frame-query similarity distribution, sampled frames, and the predicted segmentation masks for TSG and RefVOS. (a) illustrates an example for the referring task, while (b) presents an example for the reasoning task.}
\label{fig:qualitative_results}
\vspace{-2mm}
\end{figure}

As illustrated in Fig.~\ref{fig:qualitative_results}, we show qualitative results of MomentSeg on TSG and RefVOS tasks.  
For the referring task (Fig.~\ref{fig:qualitative_results}(a)), the model accurately localizes the described action and, through dense moment sampling with MCS, delivers precise segmentation.  
For the reasoning task (Fig.~\ref{fig:qualitative_results}(b)), MomentSeg identifies the temporal segment aligned with complex relational descriptions, highlighting its strong temporal reasoning.  
Additional visualizations are provided in Appendix~\ref{app_subsec:additional_qualitative_results}.

\section{Conclusion}
In this paper, we introduce MomentSeg, a unified framework that optimizes both TSG and RefVOS tasks simultaneously. Our approach eliminates the reliance on external models for keyframe selection, addressing the inherent complexity of previous methods. By leveraging a \texttt{[FIND]} token during training, MomentSeg identifies text-relevant key moments directly, bypassing the need for explicit timestamp encodings and enhancing temporal understanding. For inference, we propose Moment-Centric Sampling (MCS), which intelligently selects relevant frames while preserving essential motion cues and global temporal context. Additionally, our Bidirectional Anchor-updated Propagation (BAP) further improves tracking robustness by mitigating cumulative errors and handling occlusions effectively. Experiments on several RefVOS and TSG benchmarks demonstrate that MomentSeg outperforms existing methods, achieving SOTA results.

% \clearpage

\section{Ethics Statement}
This work focuses on algorithmic analysis and enhancements to improve the capabilities of LMMs for the referring video object segmentation (RefVOS) task.  
All datasets employed are publicly available and have undergone appropriate ethical review, ensuring lawful and responsible use.

\section{Reproducibility Statement}
To support reproducibility, we provide detailed implementation information and materials.  
Section~\ref{subsec:implementation_details} describes the construction of the baseline models, while Appendix~\ref{app_subsec:implementation_details} presents training configurations, datasets, and evaluation metrics.  
Appendix~\ref{app_subsubsec:instruction_details} further details the templates and procedures used to generate training data instructions.  
The complete codebase and trained models will be released upon publication to facilitate community verification and replication.

\bibliography{main}
\bibliographystyle{iclr2026_conference}

\appendix

\newpage

\section*{Appendix}
We provide an overview of the Appendix below. Each section includes a brief description of its contents.

\begin{itemize}[leftmargin=1.6em,itemsep=4pt]

   \item \textbf{Appendix~\ref{app_sec:llms_usage} — Use of Large Language Models (LLMs)}  

  % ========== Implementation Details ==========
  \item \textbf{Appendix~\ref{app_subsec:implementation_details} — Additional Implementation Details}  
    \begin{itemize}[leftmargin=2em,itemsep=2pt]
      \item Appendix~\ref{app_subsubsec:dataset_and_evaluation_metrics} \textit{Dataset and Evaluation Metrics} %— Composition of datasets and evaluation protocols.
      \item Appendix~\ref{app_subsubsec:training_details} \textit{Training Details} %— Implementation settings, optimizer configurations, and hyperparameters.
    \end{itemize}

  % ========== Additional Methods ==========
  \item \textbf{Appendix~\ref{app_subsec:additional_methods} — Additional Methods}  
    \begin{itemize}[leftmargin=2em,itemsep=2pt]
      \item Appendix~\ref{app_subsubsec:ttj} \textit{Temporal Token Injecting} %— Encodes temporal dynamics via specially designed temporal tokens.
      \item Appendix~\ref{app_subsubsec:instruction_details} \textit{Instruction Details} %— Provides prompts and templates for various tasks.
    \end{itemize}

  % ========== Additional Results ==========
  \item \textbf{Appendix~\ref{app_subsec:additional_results} — Additional Results}  
    \begin{itemize}[leftmargin=2em,itemsep=2pt]
      \item Appendix~\ref{app_subsubsec:qa_results} \textit{Results on Image/Video QA Tasks} %— Comparison with existing state-of-the-art multimodal models.
      \item Appendix~\ref{app_subsubsec:non-llm-based_image-video_segmentation} \textit{Comparion Image/Video Segmentation with Non-LLM-based Methods.}
    \end{itemize}

  % ========== Additional Ablation Studies ==========
  \item \textbf{Appendix~\ref{app_subsec:additional_ablation_studies} — Additional Ablation Studies}  
    \begin{itemize}[leftmargin=2em,itemsep=2pt]
      \item Appendix~\ref{app_subsubsec:ablation_temporal_tokens} \textit{Analysis of Temporal Token Injecting} %— Impact of temporal tokens on model performance.
      \item Appendix~\ref{app_subsubsec:ablation_find_token} \textit{\texttt{[FIND]} Matching Paradigm v.s. Plain-Text Generation} %— Comparison between different TSG output strategies.
      \item Appendix~\ref{app_subsubsec:ablation_frame_numbers} \textit{Effect of Frame Numbers on RefVOS} %— Influence of varying frame counts on segmentation accuracy.
      \item Appendix~\ref{app_subsubsec:ablation_post_threshold} \textit{Effect of Post-processing Threshold for TSG} %— Optimal threshold selection for temporal localization.
    \end{itemize}

  % ========== Additional Qualitative Results ==========
  \item \textbf{Appendix~\ref{app_subsec:additional_qualitative_results} — Additional Qualitative Results}  
    \begin{itemize}[leftmargin=2em,itemsep=2pt]
      \item Appendix~\ref{app_subsubsec:ablation_visualization_comparison} \textit{Visualization Comparison with Sa2VA} %— Examples of temporal sentence grounding results.
      \item Appendix~\ref{app_subsubsec:ablation_refer_visualization} \textit{Referring Instruction Visualization} %— Qualitative examples of segmentation guided by text.
      \item Appendix~\ref{app_subsubsec:ablation_reason_visualization} \textit{Reasoning Instruction Visualization} %— Visualization of reasoning-based video understanding.
      \item Appendix~\ref{app_subsubsec:ablation_badcase_analysis} \textit{Badcase Analysis} %— Insights into common failure modes and limitations.
      \item Appendix~\ref{app_subsubsec:limitations} \textit{Limitations} %— Insights into common failure modes and limitations.
    \end{itemize}

\end{itemize}

\section{Use of Large Language Models (LLMs)} 
\label{app_sec:llms_usage}
A Large Language Model was employed solely to improve the manuscript's language, including grammar, clarity, and readability.  
No technical content, experimental design, analysis, or conclusions were produced or altered by the LLM.  
All scientific ideas, methodologies, and results remain entirely the authors' original work.

\section{Additional Implementation Details}
\label{app_subsec:implementation_details}

\subsection{Dataset and Evaluation Metrics}
\label{app_subsubsec:dataset_and_evaluation_metrics}
\noindent \textbf{Datasets:} We train our MomentSeg model on a comprehensive dataset encompassing five categories: temporal sentence grounding, image question answering (QA), video QA, image segmentation, and video segmentation. The detailed composition of the training set is provided in Table~\ref{tab:train_datasets}. Our data curation largely follows the methodology of Sa2VA~\citep{yuan2025sa2va}, with notable additions including ReasonSeg (0.2K)~\citep{lai2024lisa} and two temporal sentence grounding datasets: Charades-STA (5.3K)~\citep{gao2017tall} and ActivityNet Captions (10K)~\citep{caba2015activitynet}.

Given that the Qwen2.5-VL~\citep{Qwen2.5-VL} has already been extensively pre-trained on image and video QA data, we deliberately restrict the inclusion of such data to 665K samples from LLaVA 1.5~\citep{liu2023llavaplus} and 100K from ChatUniVi~\citep{jin2024chat}, thereby mitigating the risk of catastrophic forgetting of its foundational QA capabilities. For image-level referring segmentation, we incorporate 56K referring expression data~\citep{kazemzadeh2014referitgame, yu2016modeling} and 214K grounding conversation generation data~\citep{hanoona2023GLaMM}. For video-level referring segmentation, we utilize 5.8K existing referring VOS samples from Ref-YouTubeVOS~\citep{seo2020urvos}, MeVIS~\citep{ding2023mevis}, and ReVOS~\citep{yan2024visa}.

\noindent \textbf{Evaluation Metrics:} For RefVOS, unless specified otherwise, we use the following evaluation metrics: $\mathcal{J}$ (average Intersection over Union, IoU), $\mathcal{F}$ (boundary F-measure), and $\mathcal{J}$\&$\mathcal{F}$ (the average of $\mathcal{J}$ and $\mathcal{F}$). For TSG, we report the Intersection over Union (IoU) between the predicted and ground truth timestamps, including Recall at IoU thresholds of \{0.3, 0.5, 0.7\}, as well as the mean IoU.

\begin{table}[h]
    \centering
    \caption{Composition of datasets from different tasks used during training.}
    \vspace{-2mm}
    \resizebox{1.0\textwidth}{!}{
    \begin{tabular}{l|l}
    \toprule
    Type & Datasets \\
    \midrule
    Temporal Sentence Grounding & Charades-STA (5.3K), ActivityNet Captions (10K)\\
    Image QA & LLaVA 1.5 (665K)\\
    Image Segmentation & RefCOCO (17K), RefCOCO+ (17K), RefCOCOg (22K), Grand-f (214K), ReasonSeg (0.2K)\\
    Video QA & ChatUniVi (100K)\\
    Video Segmentation & Ref-YTVOS (3.5K), MeVIS (0.6K), ReVOS (1.7K), Ref-SAV (37K)\\
    \bottomrule
    \end{tabular}
    }
    \label{tab:train_datasets}
\end{table}

\subsection{Training Details}
\label{app_subsubsec:training_details}

 Table~\ref{tab:training_details} summarizes the training configurations. Our model is trained on 8 H20 GPUs with a per-GPU batch size of 2, requiring approximately 24 hours for the 3B model. During training, we configure the maximum temporal frames to 50, TSG frames to 60, and video frames to 5. For inference, we increase the maximum TSG frames to 100, set the TSG threshold to 0.4, and use 8 video frames. The maximum pixel resolutions are configured as follows: video inputs at $16 \times 16 \times 28 \times 28$, temporal inputs at $4 \times 4 \times 28 \times 28$, TSG inputs at $8 \times 8 \times 28 \times 28$, and image inputs at $20 \times 20 \times 28 \times 28$. We employ the AdamW optimizer with $\beta_1 = 0.9$, $\beta_2 = 0.999$, and a weight decay of 0.05. The learning rate is set to $4 \times 10^{-5}$ with a warmup ratio of 0.05.

\begin{table}[h]
\centering
\caption{Implementation details of training process.}
\resizebox{0.4\linewidth}{!}{
   \begin{tabular}{l|c}
      \toprule
      Config & Value \\
      \midrule
      \textit{max temporal frame num} & 50 \\
      \textit{max TSG frame num} & 60 \\
      \textit{video frame num}& 5 \\
      \textit{max video pixels} & 16*16*28*28 \\
      \textit{max temporal pixels} & 4*4*28*28 \\
      \textit{max TSG pixels} & 8*8*28*28 \\
      \textit{max image pixels} & 20*20*28*28 \\
      \textit{optimizer} & AdamW \\
      \textit{optimizer momentum} & $\beta_1 , \beta_2 = 0.9, 0.999$ \\
      \textit{optimizer weight decay} & 0.05 \\
      \textit{learning rate} & 4e-5 \\
      \textit{LoRA rank} & 128 \\
      \textit{batch size} & 2 \\
      \textit{warmup ratio} & 0.05 \\
      \bottomrule
   \end{tabular}
   }
   
   \label{tab:training_details}
\end{table}

\section{Additional Methods}
\label{app_subsec:additional_methods}

\subsection{Temporal Token Injecting}
\label{app_subsubsec:ttj}
Beyond the visual tokens extracted from selected video frames, we introduce temporal tokens to explicitly encode temporal dynamics. These tokens are meticulously designed to capture the relative temporal ordering of frames within the video sequence, thereby substantially bolstering the model's capacity for temporal reasoning.
Our methodology draws conceptual parallels with GLUS~\citep{lin2025glus}, which integrates a finite set of global tokens, and VideoLISA~\citep{bai2024one}, which condenses video frames into a singular token representation. A key distinction lies in that our temporal tokens are exclusively injected during the training phase. Furthermore, we leverage the Qwen2.5-VL's video encoder mode to achieve significant computational efficiency. Specifically, these temporal tokens are generated from uniformly sampled video frames. Each sampled frame undergoes compression to a highly compact input dimension, with a maximum allocation of 16 tokens per individual image. The max count of uniformly sampled frames is restricted to 50. This entire process can be formally described by Eq.~\ref{eqn:momentseg_refvos}.

\subsection{Instruction Details}
\label{app_subsubsec:instruction_details}
We list the prompts used for different tasks during training in Table~\ref{tab:task_prompt}. For the TSG task, the prompt consists of video, query, and answer components, where the video component is constructed using Qwen2.5-VL's \texttt{<|video\_pad|>} placeholder. For the RefVOS task, the prompt consists of video, image, query, and answer components. Compared to TSG, RefVOS includes an additional image component, which is constructed using Qwen2.5-VL's \texttt{<|image\_pad|>} placeholder. For the answer component, we use \texttt{[FIND]} and \texttt{[SEG]} to represent the answers for TSG and RefVOS tasks, respectively. We design multiple different prompts for each component and randomly sample them during training to improve the model's robustness.

\begin{table*}
  \centering
  \resizebox{\linewidth}{!}{
    \fcolorbox{black}{gray!10}{
    \parbox{1\linewidth}{
      \vspace{0.3em}
      
      \textbf{\textcolor{blue}{Prompts for Temporal Sentence Grounding}}
      
      \vspace{0.2em}
      \textbf{Video Prompt:}
      \begin{enumerate}[label=(\arabic*)]
        \item ``\texttt{<video>}$\backslash$n This is a low-resolution video.''
        \item ``\texttt{<video>}$\backslash$n You can view this low-resolution video for reference.''
        \item ``\texttt{<video>}$\backslash$n This is a low-resolution video for analysis.''
        \item ``\texttt{<video>}$\backslash$n Here is a low-resolution video you can refer to.''
      \end{enumerate}

      \vspace{0.2em}
      \textbf{Question Prompt:}
      \begin{enumerate}[label=(\arabic*)]
        \item ``Can you find the key frames range of the text query `\texttt{<query>}' in this video?''
        \item ``Could you identify the key frames range for the text query `\texttt{<query>}' in this low-resolution video?''
        \item ``Please locate the key frames range where the text query `\texttt{<query>}' appears in this video?''
        \item ``Can you determine the key frames range containing the text query `\texttt{<query>}' in this video?''
      \end{enumerate}

      \vspace{0.2em}
      \textbf{Answer Prompt:}
      \begin{enumerate}[label=(\arabic*)]
        \item ``It is \texttt{[FIND]}.''
        \item ``Sure, \texttt{[FIND]}.''
        \item ``Sure, it is \texttt{[FIND]}.''
        \item ``\texttt{[FIND]}.''
      \end{enumerate}

      \vspace{0.5em}
      \textbf{\textcolor{blue}{Prompts for Referring Video Object Segmentation}}
      
      \vspace{0.2em}
      \textbf{Video Prompt:}
      \begin{enumerate}[label=(\arabic*)]
        \item ``\texttt{<video>}$\backslash$n This is a low-resolution video.''
        \item ``\texttt{<video>}$\backslash$n You can view this low-resolution video for reference.''
        \item ``\texttt{<video>}$\backslash$n This is a low-resolution video for analysis.''
        \item ``\texttt{<video>}$\backslash$n Here is a low-resolution video you can refer to.''
      \end{enumerate}

      \vspace{0.2em}
      \textbf{Question Prompt:}
      \begin{enumerate}[label=(\arabic*)]
        \item ``\texttt{<images>}$\backslash$n Can you segment the \texttt{<query>} in this video?''
        \item ``\texttt{<images>}$\backslash$n Please segment \texttt{<query>} in this video.''
        \item ``\texttt{<images>}$\backslash$n What is \texttt{<query>} in this video? Please respond with segmentation mask.''
        \item ``\texttt{<images>}$\backslash$n What is \texttt{<query>} in this video? Please output segmentation mask.''
        \item ``\texttt{<images>}$\backslash$n Could you provide a segmentation mask for the \texttt{<query>} in this video?''
        \item ``\texttt{<images>}$\backslash$n Please identify and segment the \texttt{<query>} in this video.''
        \item ``\texttt{<images>}$\backslash$n Where is the \texttt{<query>} in this video? Please respond with a segmentation mask.''
        \item ``\texttt{<images>}$\backslash$n Can you highlight the \texttt{<query>} in this video with a segmentation mask?''
      \end{enumerate}

      \vspace{0.2em}
      \textbf{Answer Prompt:}
      \begin{enumerate}[label=(\arabic*)]
        \item ``It is \texttt{[SEG]}.''
        \item ``Sure, \texttt{[SEG]}.''
        \item ``Sure, it is \texttt{[SEG]}.''
        \item ``Sure, the segmentation result is \texttt{[SEG]}.''
        \item ``\texttt{[SEG]}.''
      \end{enumerate}
      
      \vspace{0.3em}
    }}
  }
  \caption{Prompts used for different tasks.}
  \label{tab:task_prompt}
\end{table*}

\section{Additional Results}
\label{app_subsec:additional_results}

\subsection{Results on Image/Video QA Tasks.}
\label{app_subsubsec:qa_results}
Maintaining basic QA capabilities remains a significant challenge for LMMs designed for specific tasks. Table~\ref{tab:image_benchmarks} presents the performance of our method on image/video-based question answering tasks~\citep{fu2023mme, chen2024mmstar, kembhavi2016ai2d}. Compared to other MLLMs with segmentation capabilities, our approach achieves the best accuracy trade-off between image chat and referential segmentation datasets. Additionally, our method excels on the Video-MME~\citep{videomme} video question answering benchmark, demonstrating its strong capabilities in multimodal understanding. Furthermore, we replicate the results for Qwen2.5-VL-3B~\citep{Qwen2.5-VL} using the lmms-eval framework~\citep{zhang2024lmmsevalrealitycheckevaluation}, ensuring consistency and fairness in evaluation. After incorporating segmentation capabilities into MomentSeg based on Qwen2.5-VL, there is no significant degradation in the underlying QA performance. In fact, MomentSeg outperforms on certain evaluation sets, such as MMbench~\citep{liu2023mmbench}, SEED-Bench~\citep{li2023seed}, and MMMU~\citep{yue2024mmmu}, showcasing improvements in specific benchmarks.

\begin{table*}[t!]
    \centering
    \caption{Performance on image/video benchmarks with MLLMs possessing segmentation capabilities. The results for Qwen2.5-VL-3B~\citep{Qwen2.5-VL} are replicated using the same lmms-eval~\citep{zhang2024lmmsevalrealitycheckevaluation} framework.}
    \vspace{-2mm}
    \resizebox{\textwidth}{!}{
    \begin{tabular}{l|cccccc|c}
    \toprule
    \multirow{2}{*}{Method} &  \multicolumn{6}{c|}{Image QA} & \multicolumn{1}{c}{Video QA}\\
    \cmidrule(lr){2-7} \cmidrule(lr){8-8}
    & MME & MMBench& SEED-Bench& AI2D & MMStar & MMMU & Video-MME\\
    \midrule
    LISA-7B~\citep{lai2024lisa} & 1/1 & 0.4 & - & - & - & - & - \\
    PixelLM-7B~\citep{ren2024pixellm} & 309/135 & 17.4 & - &  - & - & - & - \\
    LaSagnA-7B~\citep{wei2024lasagna} & 0/0 & 0.0 & - &  - &  - & - & -  \\
    GLaMM-7B~\citep{hanoona2023GLaMM} & 14/9 & 36.8 & - & 28.2 & - & - & -   \\
    OMG-LLaVA-7B~\citep{zhang2024omg} & 1177/235 & 47.9 & 56.5 & 42.9 & - & - & -  \\
    Sa2VA-4B~\citep{yuan2025sa2va} & 1553/540 &  76.8 & 72.6 & 79.9 & 53.7 & 46.2  & 50.4  \\
    Sa2VA-8B~\citep{yuan2025sa2va} & \textbf{1651}/578 &  \underline{82.4} & \underline{75.5} & \underline{82.1} & \underline{60.3} & 44.7  & 52.1  \\
    Qwen2.5-VL-3B~\citep{Qwen2.5-VL} & 1511/\underline{622} & 77.6 & 74.7 & 78.4 & 55.7 & 46.4 & \underline{57.7} \\
     \midrule
    \textbf{MomentSeg-3B (Ours)} & 1491/529 & 79.2 & 74.8 & 78.3 & 56.0 & \underline{47.6} &  55.4\\
    \textbf{MomentSeg-7B (Ours)} & \underline{1595}/\textbf{684} & \textbf{83.4} & \textbf{76.7} & \textbf{82.6} & \textbf{62.1} & \textbf{50.3} & \textbf{60.9}\\
    % \textbf{MomentSeg-32B (Ours)} & & & & & & & \\
    \bottomrule
    \end{tabular}
    }
    \label{tab:image_benchmarks}
\end{table*}

\subsection{Comparison of Image/Video Segmentation with Non-LLM-based Methods}
\label{app_subsubsec:non-llm-based_image-video_segmentation}
Recent advances in small-scale models have achieved remarkable progress in traditional RIS and RVOS tasks. Several approaches~\citep{simvg,propvg} leverage powerful vision-language understanding models such as BEiT-3~\citep{beit3}, substantially enhancing referring comprehension capabilities in both image~\citep{oneref,instancevg} and video~\citep{mpg-sam2,zhang2024evf} domains. To provide a comprehensive comparative analysis, we additionally evaluate non-LLM-based methods. As shown in Table~\ref{tab:image_segmentation_nonllm}, MomentSeg, as a general-purpose video and image segmentation model, demonstrates competitive performance across diverse tasks, particularly when compared to smaller-scale models. Notably, it excels on datasets requiring temporal action understanding, such as MeVIS, where it exhibits superior capabilities in modeling temporal dynamics.

\begin{table}[t]
	\centering
   \vspace{-2mm}
	\caption{Comparison of model results with non-LLMs-based methods}
    \vspace{-2mm}
    \resizebox{1.0\textwidth}{!}{ 
	\begin{tabular}{l|ccc|ccc}
		\toprule
		% Method & RefCOCO & RefCOCO+ & RefCOCOg & GCG & ReasonSeg \\
        {Method}  & {RefCOCO} & {RefCOCO+} & {RefCOCO} & MeVIS & Ref-Youtube-VOS & Ref-DAVIS17 \\
        \midrule
        ReLA~\citep{liu2023gres}  & 73.8 & 66.0 & 65.0 & - & - & -  \\
        EEVG~\citep{eevg} & 79.5 & 71.9 & 73.6 & - & - & -  \\
        C3VG~\citep{c3vg} & 81.4 & 77.1 & 76.3 & - & - & -  \\
        OneRef-L~\citep{oneref} & 81.3 & 76.6 & 75.7 & - & - & -  \\
        DeRIS-B~\citep{deris} & 82.0 & 75.6 & 76.3 & - & - & -  \\
        \midrule
        SOC~\citep{SOC} & - & - & - & - & 67.3 & 65.8 \\
        DsHmp~\citep{DsHmp} & - & - & - & 46.4 & 67.1 & 64.9 \\
        DMVS~\citep{DMVS} & - & - & - & 48.6 & 64.3 & 65.2 \\ 
        SAMWISE~\citep{samwise} & - & - & - & 48.3 & 67.2 & 68.5 \\
        ReferDINO~\citep{referdino} & - & - & - & 49.3 & 69.3 & 68.9 \\
        MPG-SAM2~\citep{mpg-sam2} & - & - & - & 53.7 & \textbf{73.9} & 72.4 \\
        \midrule
        \textbf{MomentSeg-3B (Ours)} & \underline{82.1}  & \underline{76.9} & \underline{78.8} & \underline{54.8} & 72.0 & \underline{76.4} \\
        \textbf{MomentSeg-7B (Ours)} & \textbf{82.6} & \textbf{78.2} & \textbf{80.1} & \textbf{57.1} & \underline{72.3} & \textbf{77.4}\\ 
        % \textbf{MomentSeg-32B (Ours)} & 82.9 & 84.8 & 80.5 & 78.1 & 81.9 & 73.2 & 80.6 & 80.4 & 64.4 & 68.1 & \\
		\bottomrule
	\end{tabular}
    }
	\label{tab:image_segmentation_nonllm}
   \vspace{-4mm}
\end{table}

\section{Additional Ablation Studies}
\label{app_subsec:additional_ablation_studies}

\subsection{Analysis of Temporal Tokens Injecting.}
\label{app_subsubsec:ablation_temporal_tokens}

\begin{figure}[h]
  \centering
  \begin{minipage}{0.49\textwidth}
    \centering
    \includegraphics[width=\linewidth]{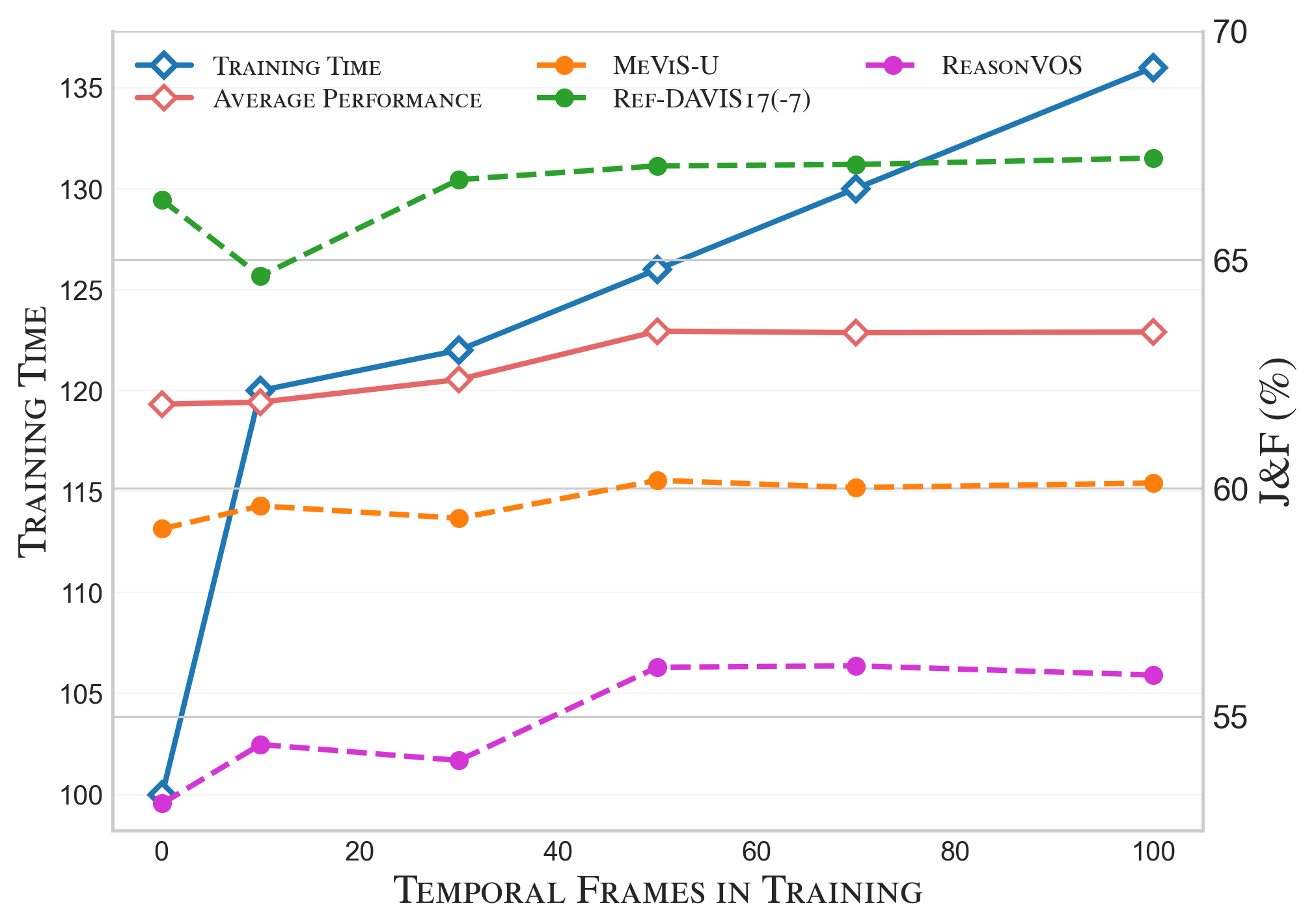}
    \caption{Effect of temporal numbers for training.}
    \label{fig:ab_videoframes_training}
  \end{minipage}
  \hfill
  \begin{minipage}{0.49\textwidth}
    \centering
    \includegraphics[width=\linewidth]{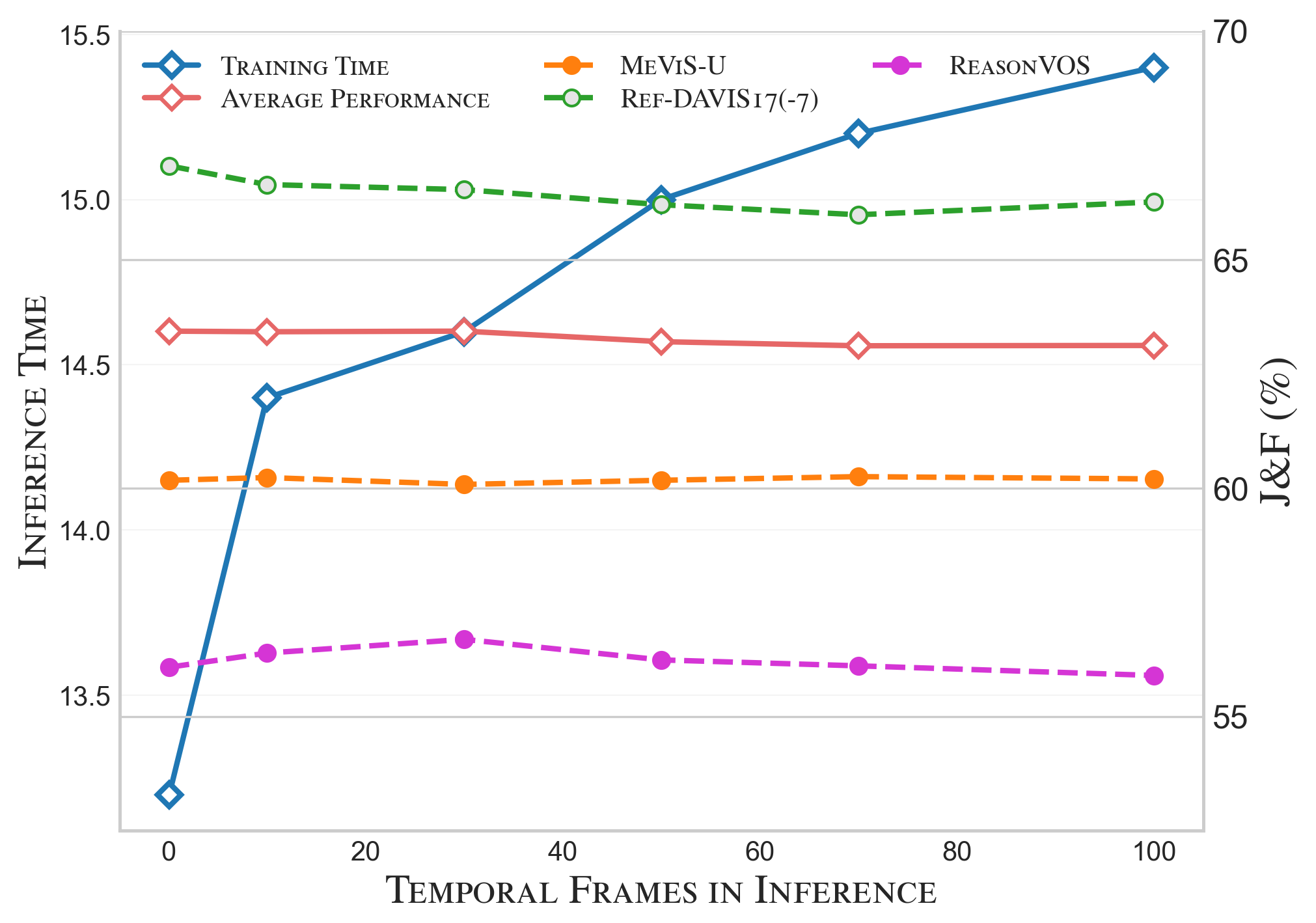}
    \caption{Effect of temporal numbers for inference.}
    \label{fig:ab_videoframes_inference}
  \end{minipage}
\end{figure}

To examine the effect of temporal tokens on model performance, we conduct ablation studies by varying their number during both training and inference.  
As shown in Fig.~\ref{fig:ab_videoframes_training}, we report results on three RefVOS datasets: MeVIS, Ref-DAVIS17, and ReasonVOS.  
During training, increasing the number of temporal frames yields a steady performance gain on MeVIS, saturating around 50 frames.  
Training cost grows proportionally with the number of temporal frames; thus we set it to 50 in all experiments.  
At inference, as also illustrated in Fig.~\ref{fig:ab_videoframes_inference}, adding temporal frames has minimal impact on accuracy.  
To improve efficiency, we therefore omit temporal frames during inference.

\subsection{\texttt{[FIND]} Matching Paradigm v.s. Plain-Text Generation}
\label{app_subsubsec:ablation_find_token}
To compare the effectiveness of the \texttt{[FIND]} token approach, we conducted an ablation experiment that simultaneously supports both matching paradigm and plain-text generation. In prompt construction, we additionally include \texttt{output with format of (start\_ratio, end\_ratio) as a 0-1 range.} after the \texttt{[FIND]} instruction. The answer construction further adds \texttt{The range is (\{\},\{\})} based on the \texttt{[FIND]} token, enabling the model to perform TSG tasks through both methods simultaneously. 
The experimental results in Table~\ref{table:app_effect_of_find} demonstrate that the \texttt{[FIND]} token approach significantly outperforms the plain-text generation method across all metrics on both Charades-STA and ActivityNet-Grounding datasets. On Charades-STA, the \texttt{[FIND]} token achieves improvements of 10.4, 11.2, and 6.5 points in R@0.3, R@0.5, and R@0.7, respectively, with a substantial 9.3-point gain in mIoU. Similarly, on ActivityNet-Grounding, the \texttt{[FIND]} token shows consistent improvements, particularly in R@0.3 and R@0.5 metrics.
The superior performance of the \texttt{[FIND]} token can be attributed to its direct prediction mechanism. Unlike the plain-text generation approach, which requires the model to understand and generate explicit timestamp encoding information, the \texttt{[FIND]} token directly enables the model to predict text-relevant key positions through similarity-based localization. This approach significantly simplifies the learning difficulty by eliminating the need for complex temporal coordinate regression, allowing the model to focus on semantic alignment between textual queries and visual content rather than numerical timestamp generation.

\begin{table}
   \centering
   \caption{Effect of post-process threshold $\theta$ on TSG.}
   \resizebox{0.8\textwidth}{!}{
      \begin{tabular}{l|ccc|c|ccc|c}
      \toprule
         \multirow{2}{*}{Method} &\multicolumn{4}{|c}{Charades-STA} &\multicolumn{4}{|c}{ActivityNet-Grounding}\\
         \cmidrule(lr){2-5} \cmidrule(lr){6-9}
         &R@0.3 &R@0.5 &R@0.7 &mIoU &R@0.3 &R@0.5 &R@0.7 &mIoU \\
            \midrule
            Plain-Text Generation   & 64.9 & 45.4 & 25.8 & 41.3 & 65.1 & 48.6 & 27.9 & 46.0 \\
            \texttt{[FIND]} Matching        & 75.3 & 56.6 & 32.3 & 50.6 & 70.3 & 52.0& 29.0 & 49.3 \\
            \bottomrule
      \end{tabular}
   }
   \label{table:app_effect_of_find}
\end{table}

\subsection{Effect of Frame Numbers on RefVOS}
\label{app_subsubsec:ablation_frame_numbers}
In the RefVOS task, the number of sampled frames strongly affects the model’s ability to identify the referred target.  
Table~\ref{table:app_ab_frame_numbers} reports performance on three RefVOS datasets under different frame counts.  
On MeViS, accuracy improves as the number of frames increases, reflecting its motion-driven nature where richer temporal context aids understanding.  
In contrast, results on Ref-DAVIS17 and ReasonVOS fluctuate, and excessive frames can even reduce accuracy.  
A moderate increase in frames can help, but too many introduce irrelevant segments that degrade segmentation.  
Moreover, the gap between training and testing—training uses only five frames—reduces the proportion of tokenized text instructions at inference, causing the model to underutilize crucial textual cues.  
This mismatch warrants further investigation.

\begin{table}
\centering
\caption{Effect of frame number $N_f$ on RefVOS. Evaluated with $\mathcal{J}$\&$\mathcal{F}$ metric.}
\resizebox{0.5\textwidth}{!}{
    \begin{tabular}{l|c|c|c}
        \toprule
        $N_f$  & {MeViS ($val^u$)} & {Ref-DAVIS17} & {ReasonVOS}   \\
        \midrule
        5  & 61.4 & 77.1 & 59.2 \\
        8  & 62.0 & 76.3 & 61.7 \\
        10 & 61.9 & 76.4 & 60.3 \\
        12 & 62.5 & 76.8 & 59.9 \\
        15 & 62.6 & 76.0 & 59.7 \\
        \bottomrule
    \end{tabular}
}
\label{table:app_ab_frame_numbers}
\end{table}

\subsection{Effect of Post-processing Threshold for TSG}
\label{app_subsubsec:ablation_post_threshold}
For the proposed MomentSeg, the post-processing threshold $\theta$ plays a crucial role in determining the accuracy of moment localization. Table~\ref{table:app_ab_post_threshold} demonstrates the impact of varying $\theta$ values on the performance of two TSG benchmarks, Charades-STA and ActivityNet-Grounding, evaluated using recall at different IoU thresholds (R@0.3, R@0.5, and R@0.7) and mean IoU (mIoU). We observe that for Charades-STA, increasing $\theta$ from 0.2 to 0.3 results in significant improvements across all metrics, particularly in R@0.3 and mIoU, where the best performance is achieved at $\theta = 0.3$. Further increasing $\theta$ leads to a decline in some lower-precision metrics, while higher-precision metrics continue to improve, indicating that a higher threshold favors retaining high-confidence moments, resulting in a high-precision, low-recall state.
Similarly, for ActivityNet-Grounding, we observe that $\theta = 0.4$ strikes a balance between recall and precision, providing an overall stable performance across the metrics. Therefore, $\theta = 0.4$ is selected as the optimal threshold for this paper.

\begin{table}
  \centering
  \caption{Effect of post-process threshold $\theta$ on TSG.}
  \resizebox{0.7\textwidth}{!}{
      \begin{tabular}{l|ccc|c|ccc|c}
      \toprule
          \multirow{2}{*}{$\theta$} &\multicolumn{4}{|c}{Charades-STA} &\multicolumn{4}{|c}{ActivityNet-Grounding}\\
          \cmidrule(lr){2-5} \cmidrule(lr){6-9}
          &R@0.3 &R@0.5 &R@0.7 &mIoU &R@0.3 &R@0.5 &R@0.7 &mIoU \\
          \midrule
          0.2  & 75.4 & 49.8 & 21.9 & 47.5 & 66.1 & 41.0 & 20.9 & 44.9 \\
          0.3  & \textbf{78.9} & 55.9 & 25.3 & \textbf{50.7} & 65.6 & \textbf{45.6} & \textbf{23.8} & 45.1 \\
          0.4 &  76.1 & \textbf{58.2} & 25.8 & 50.2& \textbf{67.5} & 44.7 &23.2 & \textbf{45.4}\\
          0.5 & 72.9 & 55.4 &  \textbf{31.1} & 49.7 & 62.2 & 42.1 & 22.4 & 42.7 \\
          0.6 & 71.8 & 54.6 & 27.9 & 48.4 & 59.2 & 39.5 & 21.7 & 41.2 \\
          \bottomrule
      \end{tabular}
  }
  \label{table:app_ab_post_threshold}
\end{table}

\section{Additional Qualitative Results}
\label{app_subsec:additional_qualitative_results}

\subsection{Visualization Comparison with Sa2VA}  
\label{app_subsubsec:ablation_visualization_comparison}
\begin{figure}[h]
\centering
\includegraphics[width=\linewidth]{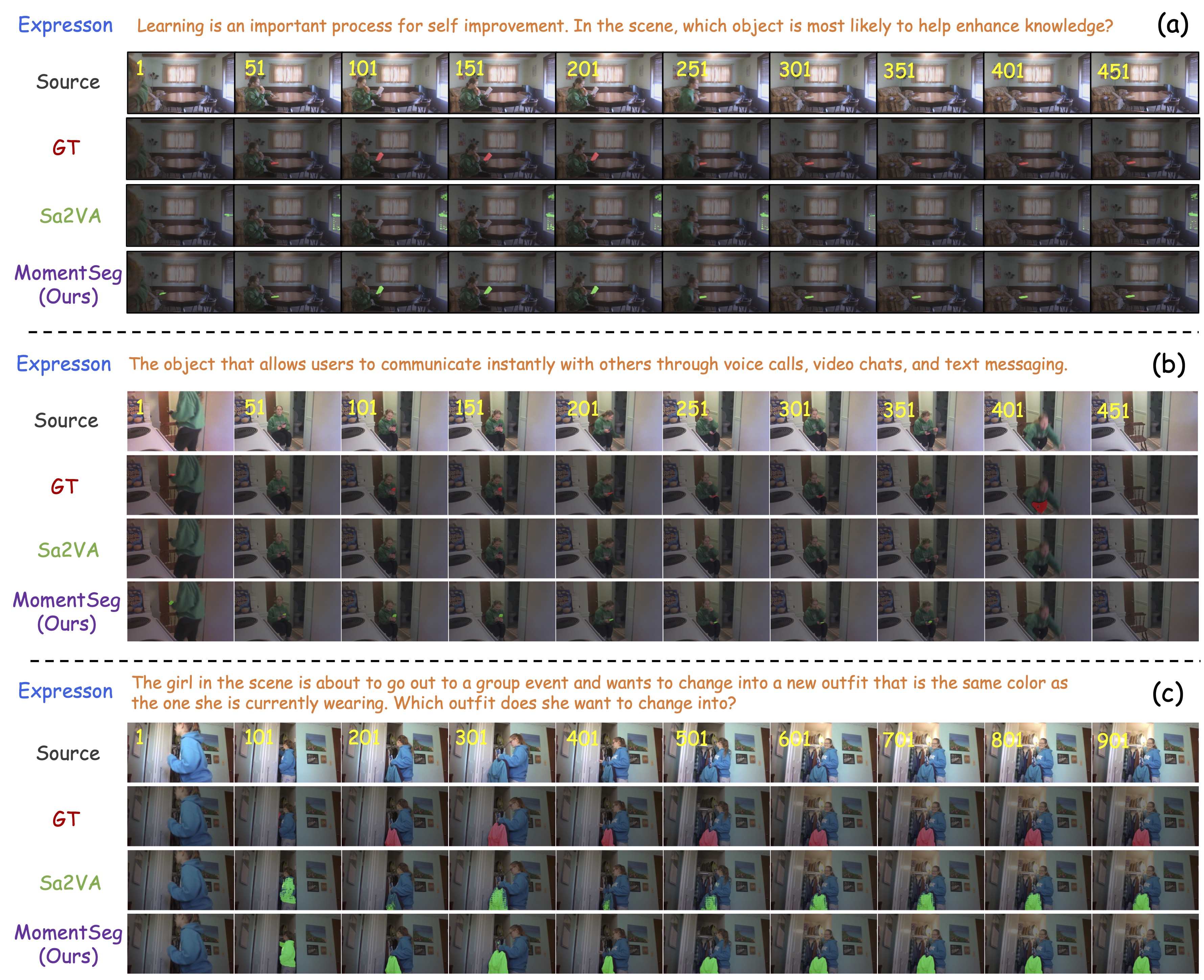}
\caption{Additional visual comparison of MomentSeg and Sa2VA on the RefVOS task.}
\label{fig:supp_comparsion_visualization}
\end{figure}

We compare the visual results of our proposed MomentSeg with Sa2VA on the RefVOS task.  
As shown in Fig.~\ref{fig:supp_comparsion_visualization}, three representative examples are provided.  
(a) Highlights a key drawback of Sa2VA’s \textit{FirstK} sampling strategy: sampling only the early portion of a video fails to capture later object appearances, leading to missed segmentations. In contrast, our TSG module accurately localizes the action described in the text and performs dense sampling within that segment, enabling correct object segmentation.  
(b) Shows a case where Sa2VA entirely misses the target. The object is small and requires precise sampling; early sampling prevents Sa2VA from observing the later `browsing phone' action.   
(c) Demonstrates that insufficient sampling of later key moments produces unreliable mask initialization and incomplete predictions for Sa2VA. Our method instead initializes at the most relevant frames and applies bidirectional propagation and updating, yielding more stable mask segmentation.

\subsection{Referring Instruction Visualization}  
\label{app_subsubsec:ablation_refer_visualization}
\begin{figure}[h]
\centering
\includegraphics[width=\linewidth]{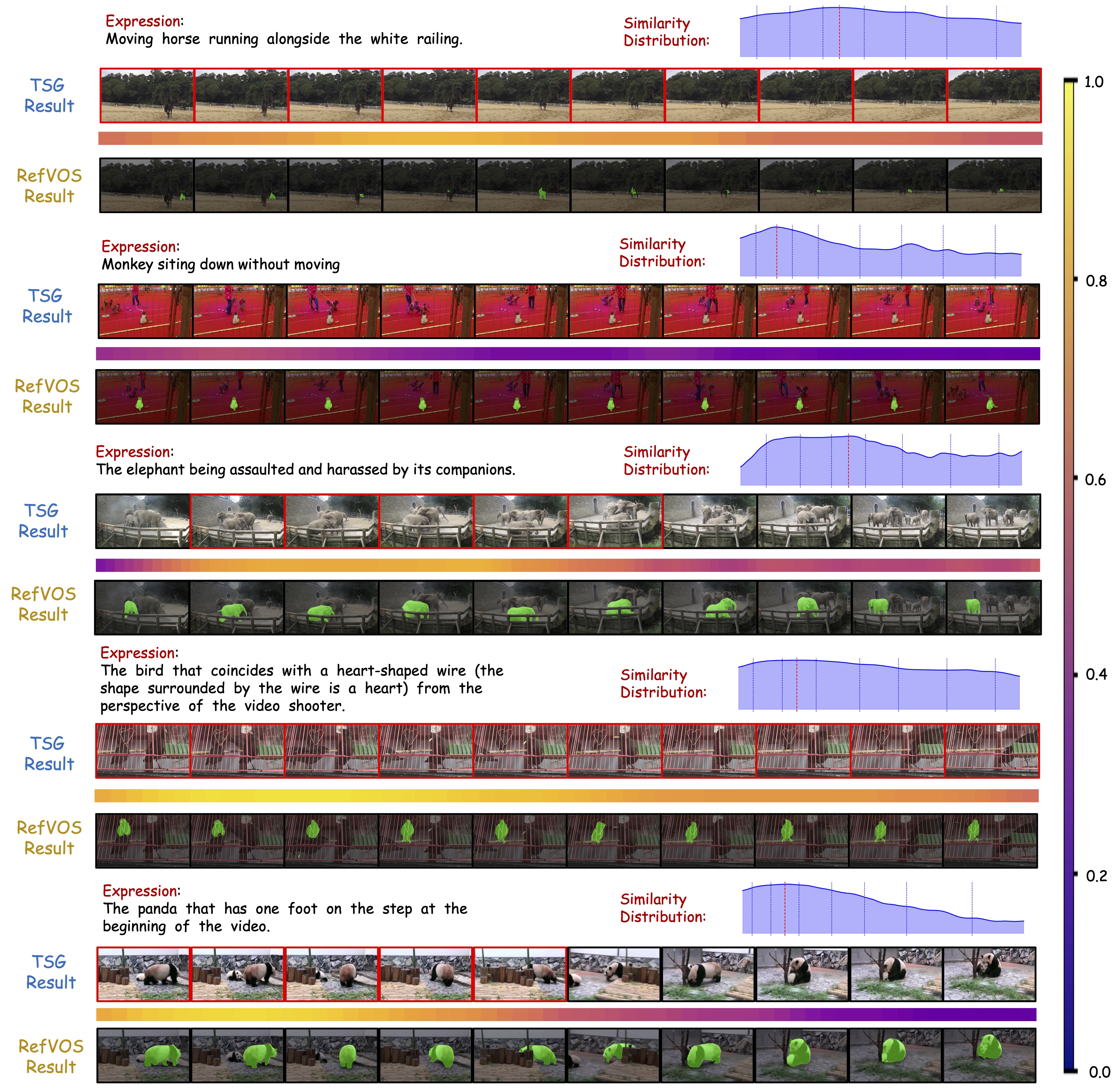}
\caption{The additional visualization of the MomentSeg on Referring VOS task.}
\label{fig:supp_refer_visualization}
\end{figure}

In Fig.~\ref{fig:supp_refer_visualization}, we present additional visualization results of MomentSeg-3B on the Referring VOS task. First, it is observed that the TSG inference stage accurately localizes the action segments described by the text. Moreover, dense sampling occurs in regions with sharp peaks in the similarity distribution, as shown in the 2nd and 5th samples. This sampling strategy effectively aids the model in understanding the action described by the text and allows for accurate tracking of the referred target during the subsequent BAP tracking process. Additionally, in complex scenarios, such as the 2nd, 3rd, and 4th samples, where multiple similar targets appear in the video, the model can still accurately segment the referred target by combining the understanding of the text description with action information.

\subsection{Reasoning Instruction Visualization}  
\label{app_subsubsec:ablation_reason_visualization}
\begin{figure}[h]
\centering
\includegraphics[width=\linewidth]{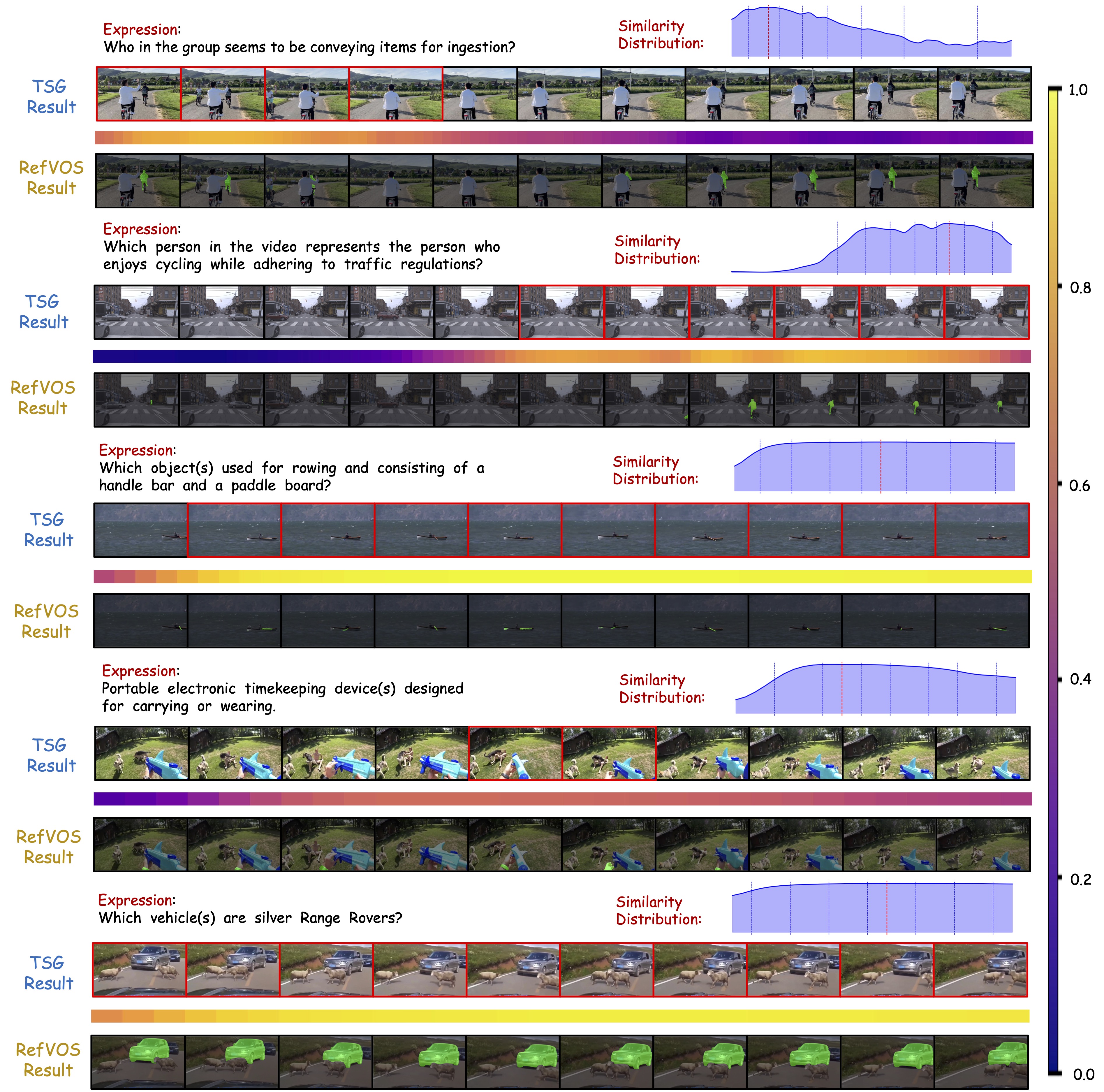}
\caption{The additional visualization of the MomentSeg on Reasoning VOS task.}
\label{fig:supp_reason_visualization}
\end{figure}

In Fig.~\ref{fig:supp_reason_visualization}, we show additional visualization results of MomentSeg-3B on the Reasoning VOS task. These examples are characterized by text descriptions that do not directly refer to a specific object but require indirect reasoning to identify the referred object. Most of these examples are in the form of questions. During this stage, it is observed that since the text does not describe much motion-related information, the TSG typically localizes the target's segment in the video rather than focusing on the critical moment. This remains a bottleneck for current methods, as we use reasoning-related TSG data for training.

\subsection{Badcase Analysis}  
\label{app_subsubsec:ablation_badcase_analysis}
\begin{figure}[h]
\centering
\includegraphics[width=\linewidth]{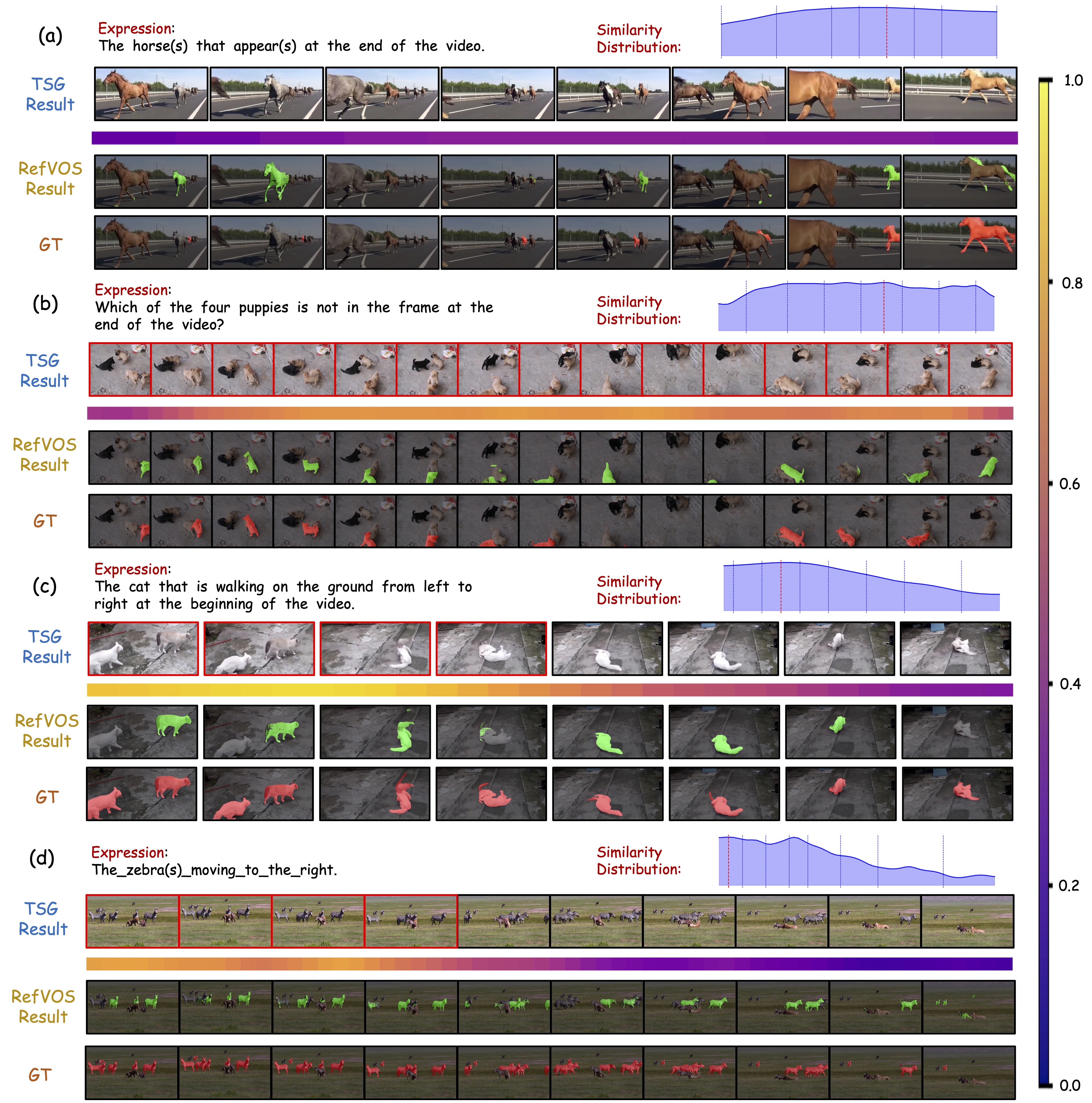}
\caption{Additional visualization of MomentSeg failure cases.}
\label{fig:supp_badcase_analysis}
\end{figure}

In Fig.~\ref{fig:supp_badcase_analysis}, we show some failure cases of MomentSeg-3B on the RefVOS task. (a) shows that due to incorrect localization of key segments during the TSG stage, the sampled frames are not representative, and with interference from multiple targets, the model misidentifies the target as smaller, resulting in erroneous and fluctuating segmentation. (b) highlights a significant issue with current methods: the lack of encoding for frame-relative video timestamps prevents the model from understanding concepts like "at the end of the video." This leads to tracking drift and inaccurate segmentation results. (c) and (d) show cases with multiple referred objects where, due to the absence of instance-level supervision, the model performs poorly. This remains a major bottleneck in current methods.

\subsection{Limitations} 
\label{app_subsubsec:limitations}
MomentSeg still offers substantial room for improvement.  
For example, for temporal grounding, the current training data remain limited and are confined to the TSG task.  
Incorporating related tasks—such as Dense Video Captioning~\citep{iashin2020multi}, Video Highlight Detection~\citep{highlights}, and Temporally Grounded Video Question Answering~\citep{chen2024cg,etbench}—could enhance the model’s generalization ability.  
Moreover, the number of sampled frames is currently fixed; future work could explore dynamically adapting the sampling rate to different video complexities, allowing more frames to assist understanding in scenes with intricate descriptions and thereby enabling more accurate target segmentation.

\end{document}